\documentclass[sigconf, nonacm]{acmart}

\DeclareUnicodeCharacter{2212}{\ensuremath{-}}
\usepackage[a-2b]{pdfx}
\usepackage{hyperref} 
\usepackage{threeparttable}
\usepackage[utf8]{inputenc}
\hypersetup{pdfencoding=unicode}
\inputencoding{utf8}
\makeatother
\usepackage{booktabs}
\usepackage{natbib}
\usepackage{graphicx}
\usepackage{amsthm}
\usepackage{amsfonts} 
\usepackage{enumitem}

\usepackage{comment}
\usepackage{subcaption}
\usepackage{color}
\usepackage[labelfont=bf,textfont=normalfont,singlelinecheck=off,
justification=centering]{subcaption}
\usepackage[ruled,vlined]{algorithm2e}
\usepackage{siunitx}
\usepackage{multirow}

\usepackage[group-separator={,},group-minimum-digits=4]{siunitx}
\usepackage{tikz}
\usetikzlibrary{shapes, arrows.meta, positioning, fit, backgrounds, shadows}

\usepackage{booktabs}
\usepackage{tabularx}

\begin{document}
\title{Real-World End-to-End Product Entity Resolution \vspace{0.4cm}}

\author{Dimitrios Karapiperis}
\orcid{0000-0002-3878-5988}
\affiliation{%
  \institution{International Hellenic University}
   \city{Thessaloniki}
  \country{Greece}
}
\email{dkarapiperis@ihu.edu.gr}

\author{George Papadakis}
\orcid{0000-0002-7298-9431}
\affiliation{%
  \institution{National and Kapodistrian University of Athens}
  \city{Athens}
  \country{Greece}
}
\email{gpapadis@di.uoa.gr}

\author{Manolis Koubarakis}
\orcid{0000-0002-1954-8338}
\affiliation{%
  \institution{National and Kapodistrian University of Athens}
  \city{Athens}
  \country{Greece}
}
\email{koubarak@di.uoa.gr}

\newcommand{\dv}[1]{\textcolor{blue}{#1}\PackageWarning{DV:}{#1!}}

\begin{abstract}
Product Entity Resolution (PER) in e-commerce is challenging due to unstructured text, formatting inconsistencies, promotional noise, and fine-grained product differences. We present an end-to-end evaluation of four zero-shot PER pipelines—--Lexical Cascade, Semantic Cascade, Hybrid Fusion Cascade, and Generative-Lexical Cascade—--across established benchmarks and introduce two real-world datasets characterized by cross-lingual noise and dense brand collisions. We evaluate lexical and dense blocking with cross-encoder, tensor-search, and Large Language Model verification, considering accuracy, runtime, and memory consumption. Our results show that lexical and semantic retrieval exhibit complementary failure modes: lexical retrieval is efficient and preserves fine-grained distinctions, while semantic retrieval is more robust to vocabulary variation but produces more false positives. Hybrid Fusion consistently achieves the highest or near-highest blocking recall. Cross-encoders provide a strong accuracy–efficiency trade-off, whereas tensor search incurs substantial precision losses. LLM verification can improve accuracy but at considerably higher computational costs. These results highlight the accuracy, latency, and infrastructure trade-offs that govern practical PER deployment.
\end{abstract}

\maketitle

\section{Introduction}

Entity Resolution (ER) is the fundamental data integration task of identifying and linking disparate records that refer to the same real-world entity~\cite{christ_survey, pap_survey}. While traditional ER frameworks have achieved remarkable success across well-structured domains such as bibliographic databases and census records, the e-commerce landscape presents a uniquely hostile environment for data integration. Product ER (PER), also referred to as \textit{Product Matching},~\cite{wdc} seeks to identify highly heterogeneous retail listings (offers) across multiple sources that refer to the identical underlying real-world catalog product. PER deviates from standard ER paradigms due to several distinct domain nuances:
\begin{itemize}[leftmargin=*]

\item Unlike structured relational tables, e-commerce listings are often scraped directly from diverse vendor user interfaces. Consequently, critical specifications (e.g., brand, model, capacity) are frequently embedded within noisy, unstructured text blocks rather than neatly aligned columns.

\item Retailers routinely employ informal aliases, aggressive abbreviations, and promotional jargon to optimize search engine visibility. A single abstract product (e.g., \textit{Sony PlayStation 5''}) may manifest as \textit{PS5 Digital Ed. Console''} on one platform and \textit{Sony Play Station Five w/ Controller''} on another. \item In standard ER (e.g., matching human names), a minor edit distance error (e.g., \textit{Jon''} vs \textit{John''}) rarely alters the entity's identity. Conversely, in PER, microscopic alphanumeric deviations dictating specifications---such as \textit{128GB''} vs \textit{256GB''}, or \textit{v1''} versus \textit{v2''}---strictly define entirely distinct entities. This sensitivity renders classical string-similarity metrics highly vulnerable to false positives~\cite{mudgal2018deep, ditto}. 

\item In e-commerce architectures, a strict conceptual boundary exists between an abstract catalog product and a vendor-specific offer. Pipelines must successfully normalize wildly varying offer descriptions (which include localized noise like \textit{``ships individually''}) to isolate the core semantic identity of the product itself.
\end{itemize}

To navigate these severe structural and lexical inconsistencies, modern PER commonly adopts a two-stage architecture designed to balance computational scalability with classification accuracy. Evaluating the strict Cartesian product of two massive product catalogs is computationally prohibitive, scaling quadratically. To bypass this bottleneck, PER decomposes the integration problem into discrete candidate generation (blocking) and entity verification (\textit{matching}) phases. The initial blocking phase aggressively prunes obvious non-matches to reduces the massive global search space down to a tightly constrained pool of highly plausible candidate pairs, ensuring that the system remains computationally tractable even when processing millions of retail listings. Subsequently, the candidate pairs that survive the blocking filter are forwarded to the matching phase, where high-precision classifiers—such as PLMs—execute rigorous, pairwise evaluations to verify real-world equivalence.


Historically, PER relied heavily on discrete lexical algorithms, including token blocking~\cite{token1}, MinHash blocking~\cite{tkde2024}, techniques leveraging BM25~\cite{sparkly}, and pBlocking~\cite{pblock}. While computationally efficient and highly precise for clean datasets, lexical pipelines are fundamentally brittle when applied to the unstructured noise of product catalogs. Because they strictly depend on vocabulary alignment, these systems inherently suffer from the lexical cascade bottleneck; true matching records that exhibit disjoint vocabularies, localized acronyms, or severe typographical errors are frequently assigned a zero-similarity score and irrevocably discarded during the initial blocking phase, establishing an absolute ceiling on the maximum achievable recall of the entire pipeline.

The integration of Transformer-based Pre-trained Language Models (PLMs), such as BERT~\cite{bert} and RoBERTa,~\cite{roberta} has fundamentally revolutionized large-scale PER. By leveraging deep semantic representations learned from massive text corpora, PLMs can successfully bridge the semantic-lexical divide, resolving complex aliases, aggressive abbreviations, and heterogeneous structural layouts that traditional lexical approaches fail to capture.   

More recently, the emergence of Large Language Models (LLMs) has introduced new paradigms for both matching and blocking. Foundational studies deploy LLMs as direct zero-shot pairwise verifiers~\cite{narayan2022can, peeters_llms}, utilizing batch-prompting~\cite{batcher} and listwise formulations~\cite{wang2025investigation} to mitigate massive auto-regressive inference costs. Alternatively, models can be shifted to an offline data ingestion phase to perform \textit{pointwise generation}~\cite{narayan2022can, wang2025investigation}. By extracting and normalizing noisy attributes into canonical representations prior to blocking, this approach achieves semantic-level recall using high-speed lexical retrieval, though it introduces permanent hallucination risks and demands massive upfront GPU allocation~\cite{batcher}.

Ultimately, while the adoption of pre-trained and generative language models has pushed the boundaries of PER, their peak performance is predominantly driven by extensive fine-tuning methodologies~\cite{karapiperis2026impact, ditto, jointbert, wdc, steiner2025fine}. Despite their robust empirical success on benchmark datasets, relying on fine-tuning introduces severe operational bottlenecks. In real-world data integration scenarios, fine-tuning is often entirely unfeasible, as it fundamentally necessitates thousands of highly accurate, task-specific labeled examples. Generating this volume of ground-truth data is prohibitively expensive and labor-intensive, making fine-tuned architectures largely impractical for low-resource environments or highly dynamic enterprise settings where domain shifts frequently render previously tuned weights obsolete~\cite{dader}. Furthermore, heavily fine-tuned PLMs typically lack robust generalization; their classification accuracy frequently degrades when they encounter out-of-distribution or unseen entities that were absent from the initial training corpus.

Despite the immense representational power of modern matching architectures, existing experimental evaluations of PER frequently fail to provide a holistic assessment across multiple structural dimensions. For example, while the benchmark introduced in~\cite{wdc} extensively evaluates state-of-the-art fine-tuned PLMs—--including those utilizing supervised contrastive pre-training followed by cross-entropy fine-tuning—--these models fundamentally lack zero-shot robustness. Consequently, they suffer from severe performance degradation when exposed to unseen entities outside their training distribution.

Similarly, the experimental evaluation by Zeakis et al.~\cite{zeakis1} is limited to analyzing the performance of various PLMs across several benchmark datasets, specifically focusing on their computational vectorization overhead, blocking scalability, and matching accuracy. However, by restricting their analysis entirely to PLMs, their evaluation framework fails to account for the emerging capabilities of zero-shot generative LLMs or the infrastructural trade-offs of hybrid lexical-semantic retrieval systems.

To address the fine-tuning bottleneck, Peeters et al. in~\cite{peeters_llms} demonstrate that zero-shot LLMs can bypass the need for task-specific training data. However, their experimental design suffers from significant methodological limitations: they evaluate LLMs strictly as pairwise verifiers during matching on pre-filtered candidate pairs. By completely ignoring the end-to-end blocking mechanisms required to scale across unaligned product catalogs, this approach fails to assess how generative verifiers handle the massive volume of uncertain pairs generated during blocking.

Papadakis et al.~\cite{pap24} critically re-evaluate established ER benchmarks by proposing theoretical linearity and complexity metrics alongside practical matcher performance margins, revealing that many standard datasets present overly simplistic classification tasks. However, their theoretical analysis is strictly bounded to lexical similarity, quantifying benchmark difficulty exclusively through token-based similarity measures such as Cosine and Jaccard similarity.. Consequently, their evaluation framework fundamentally lacks semantic similarity measures, leaving it unable to capture or assess the geometric complexity, latent relationships, and topological overlap of entities when mapped within continuous dense vector spaces.

This experimental study bridges the gap in holistic evaluations of PER by systematically dissecting both the blocking and matching phases across lexical, semantic, and generative paradigms. Specifically, we make the following contributions: 

\begin{itemize}[leftmargin=*]
\item We systematically categorize and evaluate four distinct end-to-end PER pipelines—Lexical Cascade, Semantic Cascade, Hybrid Fusion Cascade, and Generative-Lexical Cascade. This establishes a comprehensive assessment of their trade-offs regarding vocabulary sensitivity, geometric representation density, computational latency, and hardware memory consumption without relying on task-specific fine-tuning.  
\item Hybrid search and fusion methodologies have been widely adopted in the broader information retrieval literature~\cite{chen_hybrid,luan2021sparse,wu2019,zhang2024}. To the best of our knowledge, existing ER studies have not systematically evaluated this form of dual retrieval within an end-to-end PER pipeline. To address this gap, we apply a Hybrid Fusion architecture that independently executes discrete BM25 lexical blocking and continuous bi-encoder semantic search. By unifying intermediate candidate pools via Reciprocal Rank Fusion~\cite{rrf}, this approach successfully eliminates the  failure modes of isolated retrieval methods, maximizing recall without requiring complex secondary ranking models.
\item We provide a comprehensive evaluation of tensor search matching within end-to-end PER to bridge the scalability gap between discrete and continuous retrieval. By systematically comparing its deferred MaxSim scoring mechanism against full joint attention, we reveal that late-interaction models lack the deep, non-linear cross-attention required to resolve fine-grained alphanumeric contradictions in \textit{hard negatives}, demonstrating that its moderate speedup does not compensate for severe precision degradation.
\item We introduce and publicly release RETAILERS, a novel fast-moving consumer goods dataset featuring complex cross-lingual noise, embedded promotional logic, and dense brand collisions. The benchmark is provided in a curated one-to-one clean subset (RETAILERS-C) and a many-to-many dirty variant (RETAILERS-D) to rigorously test integration system robustness against authentic e-commerce heterogeneity.  
\item We provide an empirical memory utilization analysis revealing that the primary scalability ceiling in modern PER is host CPU RAM---driven by uncompressed HNSW spatial adjacency lists during dense blocking---rather than GPU VRAM during deep cross-encoder or LLM pairwise verification.
\end{itemize}

The remainder of this paper is organized as follows: Section \ref{sec:architectures} details the four evaluated end-to-end architectures along with their foundational building blocks. Section \ref{sec:setup} outlines the experimental setup, introduces the evaluation metrics, and presents the novel RETAILERS benchmark. Section \ref{sec:results} delivers a comprehensive analysis of the empirical results alongside a detailed discussion of the operational and infrastructural trade-offs. Finally, Section \ref{sec:conclusions} concludes the study.

\section{PER Architectures}
\label{sec:architectures}
In the following, we consider two product catalogs, $A$ and $B$, where $B$ is indexed for the blocking phase and $A$ provides the query records. Formally, the PER task seeks to map these catalogs by isolating the set of true matches $M = \{(a, b) \in A \times B \mid a \equiv b\}$. Rather than evaluating the intractable Cartesian space $O(\vert{}A\vert{} \times \vert{}B\vert{})$, the initial blocking mechanism acts as a high-recall filter to isolate a computationally tractable subset of candidate pairs $C \subset A \times B$, upon which the matching functions operate.

\subsection{Lexical Cascade}
\label{sec:lexical_cascade}

The lexical cascade architecture~\cite{autoem, lsblock, peeters_llms, batcher} is an asymmetric two-stage pipeline that prioritizes computational speed. It employs a low-latency lexical retrieval algorithm—most commonly Okapi BM25~\cite{robertson2009probabilistic}—as a high-recall filter to shrink the intractable global search space down to a highly concentrated set of candidate pairs.

During the blocking phase, an inverted index is constructed over the tokenized representations of all target records in $B$. This structure maps the vocabulary directly to specific records, bypassing the need for exhaustive sequential scanning. For each query record in $A$, BM25 calculates a probabilistic affinity score based on exact token overlaps—rewarding terms that are highly specific to a document while penalizing overly common vocabulary. It strictly retains only the top-$k$ target records from $B$ that exhibit a positive score. Candidate pairs surviving this lexical filter are subsequently forwarded to a downstream matching engine for pairwise verification.

By generating candidate pairs strictly through exact-token overlap, the Lexical Cascade offers two distinct advantages:

    \begin{enumerate}[leftmargin=*]
        \item \textbf{Zero Embedding Overhead:} No VRAM or disk space is expended on storing high-dimensional dense vector embeddings.
        \item \textbf{Numerical Precision:} Lexical blocking is highly sensitive to exact identifier matches, model numbers, serial numbers, and specific numeric specifications (e.g., \textit{``128GB''} vs \textit{``256GB''}), which frequently collapse or blur inside dense vector spaces.
    \end{enumerate}

However, the pipeline cannot establish connections when relevant records lack sufficient token overlap. If a true matching pair utilizes disjoint vocabularies (e.g., \textit{``Television''} vs \textit{``TV''}), BM25 assigns a score of 0, instantly dropping the candidate. Furthermore, even when partial overlap exists, typographical errors, localized acronyms, and severe format modifications radically decay the retrieval rank, frequently pushing the true match below the top-$k$ threshold before it can be evaluated by the matcher.

To illustrate the vulnerability and mechanics of the Lexical Cascade, consider two clean product records originating from separate e-commerce databases:

\begin{itemize}[leftmargin=*]
    \item \textbf{Query record $q$:} \textit{``Apple iPad Pro 11-inch 4th Gen, M2 Chip, 256GB, Space Gray''}
    \item \textbf{Candidate record $c$:} \textit{``iPad Pro 11 M2 (2022) 4G 256 GB Space Grey''}
\end{itemize}

During blocking, the tokens are extracted and normalized. While tokens like \textit{``ipad''}, \textit{``pro''}, and \textit{``256''} match perfectly, discrepancies emerge due to human writing variations: \textit{``11-inch''} versus \textit{``11''}, \textit{``4th Gen''} versus \textit{``2022''}, and \textit{``Gray''} versus \textit{``Grey''}. If the blocking phase is constrained by a restrictive top-$k$ limit, highly active competing candidates with cleaner exact token alignments—such as an older \textit{``Apple iPad Pro 11-inch 3rd Gen 256GB Space Gray''}—may outrank $c$, generating higher BM25 scores due to exact string matches on generic terms. Consequently, the downstream matcher never receives the pair $(q, c)$ to resolve its equivalence.

\subsection{Semantic Cascade}
\label{sec:semantic_cascade}

To resolve the strict vocabulary mismatch dependency inherent in the Lexical Cascade, the Semantic Cascade~\cite{jointbert, scblock, deepblocker, ditto, sudowoodo} transitions candidate pair generation from discrete token overlap to continuous metric spaces. This pipeline replaces the lexical inverted index with a dense semantic index, shifting the retrieval mechanism from exact character matching to latent geometric similarity.

During the blocking phase, a PLM-based bi-encoder~\cite{sbert} processes the records of $A$ and $B$ independently through a Siamese Transformer architecture~\cite{bert}, pooling the text of each record into a high-dimensional, continuous vector embedding. The target vectors from $B$ are then organized in a specialized vector database using Approximate Nearest Neighbor Search (ANNS) algorithms, such as HNSW~\cite{hnsw}, ScaNN~\cite{scann}, or IVF~\cite{faisslib}. When a query from $A$ is executed, the ANNS algorithm rapidly searches this geometric space—typically using cosine similarity or inner product—to retrieve the top-$k$ nearest neighbors. These semantically similar candidate pairs are subsequently forwarded to a downstream matching engine.

Transitioning to continuous vector spaces inherently exchanges lexical fragility for geometric density challenges.:

    \begin{enumerate}[leftmargin=*]
        \item The bi-encoder learns latent semantic mappings, allowing it to easily retrieve synonyms, abbreviations, and localized terms (e.g., \textit{``television''} vs \textit{``TV''}, or \textit{``PlayStation 5''} vs \textit{``PS5''}) without requiring a single matching character.
        \item Transformer-based embeddings are highly resilient to optical character recognition errors, typographical mistakes, and structural dataset heterogeneity.
    \end{enumerate}

However, this continuous space introduces significant computational overhead and geometric blind spots:
    
    \begin{enumerate}[leftmargin=*]
        \item  Dense embeddings capture macroscopic semantic meaning but often act as a low-pass filter, ignoring critical single-token numerical differences. This forces the bi-encoder to cluster \textit{highly related but distinct entities} close together, flooding the candidate set with \textit{hard negatives}.
        \item  Unlike BM25, which builds an index in seconds, building a dense index requires passing every record in $B$ through a deep neural network forward pass, costing significant GPU time.
        \item Storing high-dimensional \textit{float32} vectors for millions of records requires massive amounts of memory, necessitating complex quantization techniques (e.g., Product Quantization~\cite{pq}) to maintain operational feasibility.
    \end{enumerate}

To illustrate the power and subsequent vulnerability of the Semantic Cascade, consider the following three records:

\begin{itemize}[leftmargin=*]
    \item \textbf{Query record $q$:} \textit{``Sony PS5 Base Console, 1TB Storage''}
    \item \textbf{Candidate record $c_1$ (True Target):} \textit{``PlayStation 5 Standard Edition 1000GB''}
    \item \textbf{Candidate record $c_2$ (False Target):} \textit{``PlayStation 4 Standard Edition 1000GB''}
\end{itemize}

During blocking, a lexical BM25 blocker would completely fail to match $q$ and $c_1$ due to the zero-token overlap (\textit{``PS5''} vs \textit{``PlayStation 5''}, \textit{``1TB''} vs \textit{``1000GB''}). However, the bi-encoder recognizes the latent semantic equivalence and successfully maps the corresponding vectors to neighboring points in the vector space, capturing the true match. The inherent flaw of this architecture emerges with $c_2$. Because a \textit{``PlayStation 4''} and a \textit{``PlayStation 5''} share overwhelming semantic similarity (both are Sony gaming consoles with \textit{``1000GB''} storage), the bi-encoder maps the corresponding vectors exceptionally close to each other. Consequently, the dense blocking phase retrieves both $q$ and $c_2$ as candidates.

\subsection{Hybrid Fusion Cascade}
\label{sec:hybrid_fusion}
The Hybrid Fusion architecture includes a dual mechanism that executes both the lexical and the semantic retrieval methods. Specifically, given a query record from dataset $A$, a dual-retriever simultaneously evaluates it against dataset $B$ using two independent mechanisms: a lexical retriever utilizing BM25 blocking, and a semantic retriever utilizing a bi-encoder for ANN vector search. Each retriever independently extracts top-ranked candidate pairs up to an intermediate retrieval depth of $k^\prime$. These sets are then unified through a deterministic rank-based fusion algorithm, establishing an optimized candidate pool that is finally passed to the downstream matching engine.

Because lexical scores (probabilistic term frequencies) and semantic scores (geometric cosine similarities) exist in fundamentally incompatible mathematical distributions, they cannot be safely combined via simple linear interpolation. Instead, the architecture utilizes Reciprocal Rank Fusion (RRF)~\cite{rrf, bruch2023analysis}, a distribution-free aggregation technique detailed in Appendix~\ref{appendix:hybrid_rrf}. RRF calculates a unified score based strictly on the ordinal rank of the candidate pairs across both lists. This approach heavily penalizes entities that rank poorly in both retrievers, while aggressively rewarding entities that rank highly in either one. The fused candidates are then re-sorted by their new RRF score, and the top-$k$ unique records form the final aggregated candidate set for pairwise verification.

The Hybrid Fusion architecture aims to maximize accuracy by combining lexical and semantic search:

    \begin{enumerate}[leftmargin=*]
        \item  This architecture mitigates the complementary failure modes of the isolated retrieval pipelines. If an entity contains severe typographical errors (failing BM25), the bi-encoder catches it. If an entity has minor numeric version differences that blur in vector space (confusing the bi-encoder), BM25 successfully discriminates it.
        \item  The use of RRF eliminates the need to train complex secondary machine learning models to normalize and combine the disparate scoring mechanisms.
    \end{enumerate}
    
 However, Hybrid Fusion comes at a severe operational cost, significantly compounding  complexity and execution time:   
    \begin{enumerate}[leftmargin=*]
        \item  Because true matches are often successfully retrieved by both blocking mechanisms, the architecture wastes compute cycles executing duplicate discoveries before fusing them.
        \item Deploying this pipeline requires maintaining, updating, and synchronizing an inverted text index alongside a high-dimensional vector database. 
        \item  The total blocking latency is bounded by the slowest blocking retriever. The subsequent RRF step requires merging and re-sorting both candidate lists, adding a computational overhead proportional to the total size of the combined candidate pool.
    \end{enumerate}

To observe the synergistic effect of RRF, consider the following records:

\begin{itemize}[leftmargin=*]
    \item \textbf{Query record $q$:} \textit{``Sony PlayStation 5 Base Console, 1TB''}
    \item \textbf{Candidate record $c_1$ (False Target):} \textit{``Sony PlayStation 5 Digital Edition 1TB''}
    \item \textbf{Candidate record $c_2$ (True Target):} \textit{``PS5 Standard 1000GB''}
\end{itemize}

The lexical search aggressively matches the terms \textit{``Sony''}, \textit{``PlayStation''}, and \textit{``1TB''}, which results in retrieving $c_1$ at rank 1. However, it completely misses $c_2$ leaving it unranked. The semantic search successfully maps \textit{``PlayStation 5''} to \textit{``PS5''} and retrieves the true match $c_2$ at rank 1. 

During fusion, RRF merges the lists. Because $c_1$ scored highly in lexical but poorly in semantic, and $c_2$ scored highly in semantic but poorly in lexical, both are assigned high aggregate scores and passed into the top-$k$ fused pool. Finally, the downstream matching engine evaluates the pool. It easily identifies $c_1$ as a false positive (detecting the strict mismatch between \textit{``Base''} and \textit{``Digital''}) and correctly resolves $c_2$ as the true match—--a target that would have been entirely lost in a purely lexical cascade.

\subsection{Generative-Lexical Cascade}
\label{sec:semantic_lexical_bridge}

By decoupling semantic reasoning from the blocking phase, this architecture utilizes an LLM strictly offline to normalize the datasets into canonical representations,~\cite{narayan2022can,peeters_llms,jellyfish,brinkmann_adbis} subsequently falling back on BM25 blocking.

First, the entire raw datasets undergo a batch-processing \textit{offline} step driven by an LLM. In this capacity, the model acts as an offline normalizer, receiving strict prompts to extract core entity attributes (e.g., \textit{Brand} or \textit{Model}) from unstructured text and outputting a clean, standardized string. This offline pass maps the heterogeneous datasets to their canonical equivalents, effectively resolving synonyms, fixing typographical errors, and expanding abbreviations before any indexing occurs.

Once offline canonicalization is complete, the blocking phase mirrors the standard Lexical Cascade. Because the LLM has already forced true matching entities to share identical vocabulary (e.g., mapping both \textit{``Sony PS5''} and \textit{``Sony PlayStation Five''} to the canonical string \textit{``Sony PlayStation 5''}), the lexical blocker can recover some of the recall otherwise associated with semantic retrieval, while strictly utilizing high-speed, discrete token overlap. 

The concentrated pool of candidate pairs surviving the lexical filter is subsequently forwarded to the final verification model. Because the input text is now highly structured and devoid of localized noise, the downstream matcher experiences a significantly simplified classification landscape, allowing it to more effectively isolate true matches.

By utilizing the LLM exclusively as a pre-processor, the pipeline secures the accuracy of semantic search decreasing the traditional latency penalties:
    \begin{enumerate}[leftmargin=*]
        \item  By executing the heavy neural inference offline, blocking operates using efficient inverted-index retrieval, without requiring GPU inference at query time.
        \item  The generative model acts as a universal semantic translator, standardizing terminology across disparate databases and nullifying the primary weakness of BM25.
        \item  Unlike dense embeddings which group merely related items together, the LLM canonicalization can preserve distinctions between different versions when the relevant attributes are correctly extracted.
    \end{enumerate}
    
    However, relying on a generative model as an offline normalizer introduces massive upfront infrastructural bottlenecks and fundamentally new data integrity vulnerabilities:
    \begin{enumerate}[leftmargin=*]
        \item Executing billions of autoregressive operations over millions of database rows demands significant upfront GPU allocation. 
        \item  If the LLM generates fabricated specifications (hallucinations) during the canonicalization of a highly noisy record, the error becomes permanently encoded in the BM25 index, potentially causing a retrieval failure.
    \end{enumerate}

To illustrate the power of offline canonicalization, consider the highly heterogeneous pair:
\begin{itemize}[leftmargin=*]
    \item \textbf{Query record $q$:} \textit{``Sammy Galaxy S-Twenty Two Ultra, Black, 256 Gigs''}
    \item \textbf{Candidate record $c$:} \textit{``SM-S908U1 Samsung S22 U 256GB Phantom Blk''}
\end{itemize}

A standard BM25 blocker would score this pair at $0$ due to complete vocabulary divergence. A bi-encoder might retrieve it, but would also retrieve the \textit{``128GB''} and \textit{``S21''} variants due to semantic clustering. In the Generative-Lexical Cascade, during the offline canonicalization, the LLM recognizes \textit{``Sammy''} as an informal alias for \textit{``Samsung''}, resolves \textit{``S-Twenty Two''} as \textit{``S22''}, and maps the model number \textit{``SM-S908U1''} to its commercial name. The LLM outputs identical canonical strings for both $q$ and $c$ to \textit{``Samsung Galaxy S22 Ultra, 256GB, Black''}. During blocking, BM25 assigns this pair the maximum possible score, instantly retrieving the true match and allowing the downstream engine to easily verify the equivalence.

\subsection{Downstream Matching Engines}
\label{sec:downstream_matchers}

Once the initial blocking cascade reduces the intractable search space down to a computationally manageable candidate set, these pairs are forwarded to a downstream matching engine for rigorous pairwise verification. In modern PER pipelines, three primary architectures are utilized for this final classification phase, namely cross-encoders, tensor search, and LLMs.

The \textbf{cross-encoder}~\cite{ditto, jointbert} utilizes a PLM to act as the primary high-precision verification engine. Unlike bi-encoders, which generate separate vectors for isolated retrieval, a cross-encoder concatenates the raw text of two candidate records into a single sequence separated by control tokens. This unified sequence is processed through multi-head self-attention layers, allowing full token-level interactions and deep semantic cross-referencing between the source and target sequences. A linear classification head then projects the final pooled representation into a binary matching probability. Because this joint encoding process is computationally heavy, cross-encoders cannot be used for database-wide retrieval and are strictly reserved for this final evaluation of the pre-filtered candidate pairs.

The \textbf{tensor search} matching method utilizes an architecture based on late-interaction models like ColBERT~\cite{colbert1, colbert2, colbert3}, encoding records into matrices of contextualized token-level embeddings rather than a single pooled vector. During scoring, it captures fine-grained alignments by computing the sum of maximum cosine similarities (MaxSim) between the token tensors of two candidate records. By deploying tensor search exclusively as a downstream pairwise evaluator for pre-filtered candidates, we successfully bypass the massive memory footprint typically required for global multi-vector indexing. However, while this decoupled approach executes faster than the full self-attention of a cross-encoder, our evaluations reveal that this speedup ultimately fails to compensate for a significant degradation in overall matching precision.

Operating as a direct, zero-shot replacement for the cross-encoder, a generative \textbf{LLM}~\cite{peeters_llms, batcher, steiner2026labelingtrainingdataentity} receives the candidate pair within a strictly constrained prompt template. Instead of relying on task-specific fine-tuning or a classification head, the LLM is instructed to determine entity equivalence based entirely on its broad pre-trained semantic knowledge. The model autoregressively evaluates the pair and outputs a constrained binary token (e.g., \textit{``Yes''} or \textit{``No''}), resolving the match without requiring localized training data.

\section{Experimental Setup}
\label{sec:setup}

\textbf{Datasets.} The benchmark datasets utilized to evaluate the PER architectures focus on real-world e-commerce heterogeneity, capturing varying degrees of structural quality across evaluation domains. Specifically, these benchmarks distinguish between \textit{clean} datasets, which contain highly standardized, internally deduplicated records, and \textit{dirty} variants characterized by noisy, unstructured text and missing values, and variable matching cluster sizes. For lexical blocking with BM25, we index only concise, discriminative attributes (e.g., product titles) to prevent long descriptions from diluting exact token matches with noise. Conversely, for dense embeddings, we concatenate titles with rich descriptive fields, allowing transformer-based bi-encoders to leverage broader semantic context and capture latent similarities across the full text. Table~\ref{tab:datasets} summarizes the structural characteristics, target attributes, and scales of these evaluation benchmarks. 

\begin{table*}[tp]
\centering
\caption{PER Benchmark Datasets. $\vert{}A\vert{}$ and $\vert{}B\vert{}$ denote collection sizes, and $\vert{}M\vert{}$ the total ground-truth matches. \textit{Lexical} and \textit{Semantic} indicate the attributes used for BM25 and dense indexing, respectively, while \textit{Gemma} and \textit{GPT} report the offline canonicalization time (in seconds). }
\label{tab:datasets}
\begin{tabular}{llrrrllrr}
\toprule
\textbf{Dataset} & \textbf{ER Type} & \textbf{$|A|$} & \textbf{$|B|$} & \textbf{$|M|$} &Lexical&Semantic&Gemma&GPT \\ \midrule
ABT-BUY$^\dagger$  & Clean & 1,082 & 1,093 & 1,097&\textit{name} & \textit{name, description} 
&\num{2152.14}&145.12\\
\addlinespace
AMAZON-GOOGLE$^\dagger$ & Dirty & 1,364 & 3,227 & 1,300 & \textit{title} & \textit{title, description}&\num{4604.58}&310.49\\
\addlinespace
AMAZON-WALMART$^\dagger$  & Dirty & 2,555 & 22,075 & 1,020 & \textit{title} & \textit{title, shortdescr}&\num{15614.45}&935.56\\
\addlinespace
RETAILERS-C  & Clean & 1,491 & 1,466 & 1,459& \textit{name} & \textit{name}
\\
\addlinespace
RETAILERS-D  & Dirty & 12,023 & 12,023 & 12,023 & \textit{name} & \textit{name}
\\
\addlinespace
WDC-Products~\cite{wdc} & Dirty & 100,000 & 2,000,000 & 10,180$^*$ & \textit{brand, title} & \textit{brand, title, description}
\\
\bottomrule
\multicolumn{7}{l}{\footnotesize $^\dagger$\url{https://github.com/anhaidgroup/deepmatcher/blob/master/Datasets.md}}\\
\multicolumn{7}{l}{\footnotesize $^*$The total number of positive pairs included in the respective train, validation, and test splits.}
\end{tabular}
\end{table*}

The \textbf{ABT-BUY} dataset is a \textit{clean} collection of product names and descriptions. Although both catalogs of ABT-BUY are internally clean and deduplicated, the total ground truth matches (\num{1097}) exceed the smaller ABT dataset (\num{1082}) due to one-to-many cross-catalog mappings. This discrepancy arises because retailers assign Stock Keeping Units (SKUs) at different granularities; a single base-product SKU in one catalog can legitimately match multiple distinct variant SKUs (e.g., different colors or bundles) in the other. The \textbf{AMAZON-GOOGLE} benchmark is often employed in its \textit{dirty} variant to test robustness against noisy formats. The \textbf{AMAZON-WALMART} dataset provides another layer of structural complexity, while the \textbf{WDC-Products} dataset is a multi-dimensional large-scale benchmark extracted from over \num{3000} e-shops. 

In this paper, we introduce \textbf{RETAILERS},\footnote{We gratefully acknowledge Pockee (\url{https://pockee.com}) for providing the industrial e-commerce data utilized to construct these benchmark datasets.} a novel, real-world benchmark dataset sourced from the fast-moving consumer goods and e-commerce sector. By capturing the authentic noise of global e-commerce, RETAILERS uniquely challenges current data integration systems to handle lexical heterogeneity, detect semantic contradictions, and perform robust cross-lingual product matching. 

The benchmark consists of two unaligned product catalogs along with a manually verified ground truth mapping. Both catalogs contain unstructured, noisy product titles alongside unique platform identifiers. The RETAILERS dataset exhibits complex morphological and semantic challenges that may degrade the recall of token-based algorithms and the precision of standard dense vector embeddings:
\begin{itemize}[leftmargin=*]
\item \textbf{Cross-lingual Noise:} Product descriptions heavily mix local category descriptors with international English brand names, necessitating robust cross-lingual representation models.
\item \textbf{Inconsistent Formatting:} Measurement units are inconsistently transliterated across the catalogs (e.g., $1L$ vs. $1lt$, and $gr$ vs. $g$).
\item \textbf{Promotional Logic:} The dataset frequently embeds dynamic pricing tags and promotional logic directly into the entity string (e.g., \textit{-0.10\$} or \textit{40+20 Free}), causing false negative semantic shifts when compared to standard retail listings.
\item \textbf{Brand Collisions:} The dataset is exceptionally rich in pairs with high lexical overlap that refer to entirely different entities from competing manufacturers (e.g., \textit{Durex} vs. \textit{Duo}).
\end{itemize}

These structural characteristics generate representative edge cases that underscore the necessity for hybrid architectures. While true matches might exhibit slight lexical variations due to promotional text or minor descriptive additions (e.g., aligning \textit{``Kings Soft Cookie Triple Chocolate 160gr''} with \textit{``Kings Soft Cookie Triple Chocolate Chunks 160gr''}), the primary challenge lies in resolving severe \textit{hard negatives}. For example, distinct products like the \textit{``NIVEA SUN Protect \& Moisture Spray''} variants differing only by their sun protection factor (\textit{SPF50+} versus \textit{SPF30}), or \textit{``LIPTON Ice Tea''} variants differing merely by flavor (\textit{Peach} versus \textit{Lemon}), share almost completely identical semantic contexts. Because dense embeddings often act as a low-pass filter and ignore these critical single-token differences, they cluster these highly related but distinct entities close together, necessitating a precise verification step to successfully filter them out.

To support comprehensive evaluation and facilitate further research, we make RETAILERS publicly available in two distinct versions:
\begin{itemize}[leftmargin=*]
\item \textbf{RETAILERS-C}: A curated \textit{clean} subset focused on one-to-one matching. 
\item \textbf{RETAILERS-D}: A larger, \textit{dirty} variant containing \num{12023} products in each source ($A$ and $B$). This version reflects real-world complexities by preserving many-to-many mappings, where a single product in one source may resolve to multiple products in the other.
\end{itemize}

For the Generative-Lexical Cascade, the execution time of records canonicalization, reported in Table~\ref{tab:datasets}, scales linearly with dataset size. Because this is a one-time offline preprocessing step, its execution time is omitted from the reported runtime metrics to ensure a fair comparison of the online pairwise retrieval and verification stages. Furthermore, we omit this cascade entirely from the RETAILERS-C and RETAILERS-D evaluations, as these datasets lack the extensive textual descriptions that necessitate LLM-based canonicalization. For the WDC-Products dataset, we deliberately omitted downstream LLM matchers and generative canonicalization, as pairwise LLM prompting becomes computationally and financially prohibitive at this massive scale.

\textbf{Metrics.} We report end-to-end precision, recall, F1-score, execution runtime, and peak memory utilization for all evaluated architectures. In the context of ER, precision quantifies the quality (exactness) of the evaluated method (the fraction of predicted matches that are true duplicates), while recall measures its completeness (the fraction of all actual ground-truth matches successfully retrieved by the method). The F1-score computes the harmonic mean of these two metrics, providing a single balanced measure of overall matching quality. To evaluate infrastructural hardware requirements, we also report peak memory utilization, explicitly isolating host memory (CPU RAM) from device memory (GPU VRAM).

When measuring blocking quality, it is important to note that the evaluation metrics are inherently bounded by a dataset's matching cardinality. Consequently, when evaluating at a retrieval depth of $k$, the maximum achievable true positive count per query is exactly one for datasets with at most one ground-truth match per query record. This mathematically restricts the blocking precision ceiling to $1/k$ (e.g., a maximum of $0.2$ for $k=5$), which subsequently anchors the maximum possible F1-score.

\textbf{Implementation Details.} Lexical blocking is powered by Tantivy\footnote{\url{https://github.com/quickwit-oss/tantivy}} via its Python bindings,\footnote{\url{https://github.com/quickwit-oss/tantivy-py}} while the $384$-dimensional \texttt{BAAI/\allowbreak bge-small-en-v1.5} model~\cite{bge} is utilized to generate the dense embeddings, providing an architecture explicitly optimized via contrastive learning to enhance entity separability in dense vector spaces. Following vectorization, these dense vectors are organized within a Hierarchical Navigable Small World (HNSW) index~\cite{hnsw} implemented via the FAISS library~\cite{faisslib}. This ANNS structure enables highly efficient, sub-linear retrieval, effectively eliminating the bottleneck of exhaustive pairwise similarity calculations.

Semantic verification is performed using (a) the \texttt{ms-marco-electra\allowbreak-base} cross-encoder (Electra),\footnote{\url{https://huggingface.co/cross-encoder/ms-marco-electra-base}} (b) the \texttt{colbert-ir/colbertv2.0} late-interaction tensor search model (ColBERT),\footnote{\url{https://huggingface.co/colbert-ir/colbertv2.0}} (c) a local LLM, \texttt{Gemma-4-\allowbreak12B-it}~\cite{gemma} (Gemma), executed via the Ollama backend,\footnote{\url{https://ollama.com/}} and (d) a remote commercial LLM, \texttt{GPT-5} (GPT), accessed via API. 
Across all evaluated architectures, the final candidate pool is constrained to $k=5$, unless otherwise stated. During Hybrid Fusion, the independent retrieval pools are expanded to an intermediate depth of $k^\prime=40$ before applying RRF.

The experiments\footnote{Source code and datasets can be found at \url{https://anonymous.4open.science/r/Product-Entity-Resolution-386B/}} were conducted on a virtual server equipped with 32~GB of system RAM and a T4 GPU featuring 15~GB of VRAM. Furthermore, no domain-specific fine-tuning~\cite{karapiperis2026impact} was applied to either the bi-encoder or the cross-encoder. We intentionally deploy these models out-of-the-box to simulate realistic, zero-shot PER scenarios, as acquiring sufficient ground-truth annotations for fine-tuning is often practically unfeasible in real-world industrial applications. All reported metrics represent the empirical mean across the 10 independent runs. However, due to the high variability observed in the execution times of the LLM-based matching pipelines, we report all runtimes as the empirical mean alongside the standard deviation ($\sigma$) to accurately capture this computational variance.

\begin{table}[htbp]
    \centering
    \footnotesize
    \caption{Overview of the Evaluated PER Architectures.}
    \label{tab:method_abbreviations}
    \renewcommand{\arraystretch}{1.2}
    \begin{tabular}{@{}lp{5.8cm}@{}}
        \toprule
        \textbf{Abbr.}  & \textbf{Description} \\
        \midrule
        \multicolumn{2}{c}{\textbf{Lexical Cascades}} \\ 
        \midrule  
        \textbf{B25-B} & Lexical Blocking using the BM25 retrieval function. \\
        \textbf{B25-CM}  & Cross-Encoder Matching applied after B25-B. \\
        \textbf{BM25-TS} & Tensor Search Matching applied after B25-B. \\
        \textbf{B25-LM-L}  & Local LLM Matching applied after B25-B. \\
        \textbf{B25-LM-R} & Remote LLM Matching applied after B25-B. \\
        \midrule
        \multicolumn{2}{c}{\textbf{Semantic Cascades}} \\ 
        \midrule          
        \textbf{BE-B} & Semantic Blocking utilizing a Bi-Encoder. \\
        \textbf{BE-CM} &  Cross-Encoder Matching applied after BE-B.\\
        \textbf{BE-TS} &  Tensor Search Matching applied after BE-B.\\        
        \textbf{BE-LM-L} & Local LLM Matching applied after BE-B.\\
        \textbf{BE-LM-R} & Remote LLM Matching applied after BE-B.\\
        \midrule
        \multicolumn{2}{c}{\textbf{Hybrid Fusion Cascades}} \\ 
        \midrule          
        \textbf{HB-B}  & Hybrid Blocking using BM25 and a Bi-Encoder. \\
        \textbf{HB-CM} & Cross-Encoder Matching applied after HB-B.\\
        \textbf{HB-TS} & Tensor Search Matching applied after HB-B.\\        
        \textbf{HB-LM-L} & Local LLM Matching applied after HB-B. \\
        \textbf{HB-LM-R} & Remote LLM Matching applied after HB-B.\\
        \midrule
        \multicolumn{2}{c}{\textbf{Generative-Lexical Cascades}} \\ 
        \midrule                  
        \textbf{B25-CB-L} & BM25 Blocking over canonical entity descriptions generated by a local LLM.\\
        \textbf{B25-CCM-L} & Cross-Encoder Matching applied after B25-CB-L.\\
        \textbf{B25-CTS-L} & Tensor Search Matching applied after B25-CB-L.\\        
        \textbf{B25-CB-R} & BM25 Blocking over canonical entity descriptions generated by a remote LLM. \\
        \textbf{B25-CCM-R} & Cross-Encoder Matching applied after B25-CB-R.\\
        \textbf{B25-CTS-R} & Tensor Search Matching applied after B25-CB-R.\\        
        \bottomrule
    \end{tabular}
    \vspace{-5mm}
\end{table}

\section{Experimental Results}
\label{sec:results}
Table~\ref{tab:method_abbreviations} provides a comprehensive overview of the evaluated blocking and matching methods---the fundamental building blocks of the PER architectures detailed in Section~\ref{sec:architectures}---along with their corresponding abbreviations.

\subsection{Evaluation Across Benchmark Datasets}

For \textbf{ABT-BUY}, the baseline blocking methods (B25-B, BE-B, and HB-B) consistently achieve high recall, all exceeding $0.91$ (peaking at $0.952$ for HB-B). Introducing cross-encoder (B25-CM, BE-CM, HB-CM) or tensor search (B25-TS, BE-TS, HB-TS) matching engines provides only marginal F1-score improvements over the pure blocking baselines, keeping them anchored around $0.32$ while increasing execution times to approximately $13$ to $15$ seconds. The most significant performance leaps occur when integrating LLMs. Gemma-based matching methods deliver the highest overall effectiveness, with BE-LM-L achieving a peak F1-score of $0.872$ and a recall of $0.914$. However, this accuracy incurs a massive computational cost, requiring nearly \num{1300} seconds to execute. The GPT-based methods generally underperform their Gemma counterparts on this dataset; for example, BE-LM-R yields an F1-score of $0.579$, representing a $33.6\%$ drop in F1-score compared to BE-LM-L, though it executes faster at $770$ seconds. Finally, the Gemma canonical cross matching method (B25-CCM-L) offers a compelling middle ground, achieving a strong F1-score of $0.667$ in just $3.4$ seconds, providing a highly practical balance of efficiency and effectiveness.

\begin{table}[ht]
\centering
\small
\caption{Evaluation results for ABT-BUY.}
\label{tab:abt_buy_results_new}
\begin{tabular}{l c c c r r}
\toprule
\textbf{Method} & \textbf{Prec.} & \textbf{Rec.} & \textbf{F1} & \textbf{Time (sec)} & $\sigma$\\
\midrule
B25-B     & 0.185 & 0.912 & 0.308 & 0.17&  \\
B25-CM    & 0.188 & 0.912 & 0.312 & 15.40&  \\
B25-TS    & 0.185 & 0.912 & 0.308 & 13.65&  \\
B25-LM-L  & 0.844 & 0.868 & 0.856 & \num{1437.25}& 10.52 \\
B25-LM-R  & 0.693 & 0.576 & 0.629 & 120.26 &45.46 \\
\midrule
BE-B      & 0.193 & 0.944 & 0.320 & 4.99 &\\
BE-CM     & 0.195 & 0.944 & 0.323 & 15.09& \\
BE-TS     & 0.193 & 0.944 & 0.320 & 13.61& \\
BE-LM-L   & 0.833 & 0.914 & 0.872 & \num{1299.73}& 11.11 \\
BE-LM-R   & 0.726 & 0.481 & 0.579 & 770.25 &53.62 \\
\midrule
HB-B      & 0.192 & 0.952 & 0.320 & 7.12& \\
HB-CM     & 0.194 & 0.952 & 0.322 & 15.22& \\
HB-TS     & 0.192 & 0.952 & 0.320 & 13.66 &\\
HB-LM-L   & 0.836 & 0.906 & 0.870 & \num{1414.48} &12.16\\
HB-LM-R   & 0.674 & 0.281 & 0.396 & \num{1497.55} &87.72\\
\midrule
B25-CB-L  & 0.171 & 0.840 & 0.285 & 0.08 &\\
B25-CCM-L & 0.635 & 0.703 & 0.667 & 3.41 &\\
B25-CTS-L & 0.595 & 0.670 & 0.630 & 3.81 &\\
\midrule
B25-CB-R  & 0.156 & 0.755 & 0.258 & 0.07 &\\
B25-CCM-R & 0.303 & 0.713 & 0.425 & 3.51 &\\
B25-CTS-R & 0.309 & 0.698 & 0.429 & 3.91 &\\
\bottomrule
\end{tabular}
\end{table}

The \textbf{AMAZON-GOOGLE} benchmark (Table~\ref{tab:amazon_google_results}) proves to be a significantly more challenging dataset, characterized by a steep decline in overall matching quality and substantially longer computational times. The baseline blocking techniques achieve recall rates ranging from $0.79$ to $0.83$, but exhibit low precision (roughly $0.13$),\footnote{In dirty datasets, blocking precision is mathematically capped by each entity's true duplicate count relative to $k$.} leading to F1-scores confined to the $0.22$ to $0.23$ range. The application of cross-encoder matching (B25-CM, BE-CM, HB-CM) provides only marginal improvements, raising F1-scores slightly to approximately $0.28$. In contrast, tensor search matching (B25-TS, BE-TS, HB-TS) completely fails to elevate the baseline accuracy, yielding F1-scores near $0.23$—--an approximate $19\%$ relative degradation compared to the cross-encoder configurations. The local Gemma models deliver the highest overall matching accuracy; BE-LM-L and HB-LM-L peak at F1-scores of $0.53$ while maintaining strong recall rates near $0.79$. However, this accuracy incurs a severe computational cost, requiring over \num{2700} seconds to execute. Most notably, the GPT matching methods severely underperform. For instance, BE-LM-R yields a minimal F1-score of $0.2$ and a low recall of $0.14$. Compared to its Gemma counterpart (BE-LM-L), this represents a massive $62.3\%$ reduction in F1-score and an $82\%$ drop in recall. Interestingly, the canonical methods offer the most compelling balance of efficiency and accuracy. Specifically, B25-CCM-R achieves an F1-score of $0.43$ with a robust recall of $0.79$. While it does not quite match the $0.53$ F1-score of BE-LM-L pipeline, B25-CCM-R executes in under $4$ seconds, compared to over \num{2700} seconds required by BE-LM-L. 

\begin{table}[ht]
\centering
\small
\caption{Evaluation results for AMAZON-GOOGLE.}
\label{tab:amazon_google_results}
\begin{tabular}{l c c c r r}
\toprule
\textbf{Method} & \textbf{Prec.} & \textbf{Rec.} & \textbf{F1} & \textbf{Time (sec)} & $\sigma$ \\
\midrule
B25-B     & 0.129 & 0.793 & 0.222 & 0.67 &\\
B25-CM    & 0.170 & 0.760 & 0.278 & 57.44 &\\
B25-TS    & 0.134 & 0.723 & 0.227 & 31.93 &\\
B25-LM-L  & 0.393 & 0.756 & 0.517 & \num{2722.76} &11.62 \\
B25-LM-R  & 0.287 & 0.259 & 0.272 & 876.55 &41.76\\
\midrule
BE-B      & 0.135 & 0.824 & 0.232 & 10.64 &\\
BE-CM     & 0.175 & 0.797 & 0.287 & 57.30 &\\
BE-TS     & 0.136 & 0.731 & 0.229 & 31.71 &\\
BE-LM-L   & 0.394 & 0.794 & 0.527 & \num{2732.63} &17.36\\
BE-LM-R   & 0.324 & 0.143 & 0.198 & \num{1970.38} &88.16\\
\midrule
HB-B      & 0.134 & 0.831 & 0.231 & 11.59 &\\
HB-CM     & 0.173 & 0.789 & 0.284 & 57.51 &\\
HB-TS     & 0.134 & 0.720 & 0.226 & 31.86 &\\
HB-LM-L   & 0.393 & 0.793 & 0.525 & \num{2776.14} &16.26\\
HB-LM-R   & 0.366 & 0.163 & 0.225 & \num{1863.72} &75.12\\
\midrule
B25-CB-L  & 0.141 & 0.855 & 0.243 & 0.20 &\\
B25-CCM-L & 0.291 & 0.709 & 0.412 & 3.71 &\\
B25-CTS-L & 0.329 & 0.709 & 0.449 & 3.71 &\\
\midrule
B25-CB-R  & 0.154 & 0.929 & 0.265 & 0.20 &\\
B25-CCM-R & 0.297 & 0.788 & 0.432 & 3.67 &\\
B25-CTS-R & 0.297 & 0.759 & 0.427 & 3.87 &\\
\bottomrule
\end{tabular}
\end{table}

The \textbf{AMAZON-WALMART} (Table~\ref{tab:amazon_walmart_results}) benchmark reveals severe challenges for standard blocking and matching pipelines, characterized by exceptionally low precision across most evaluated techniques. Baseline blocking methods successfully maintain high recall rates ranging from $0.86$ to $0.89$, but suffer from extremely poor precision (approximately $0.08$). Consequently, baseline F1-scores languish around $0.14$. Introducing cross-encoder matching (B25-CM, BE-CM, HB-CM) provides minimal benefit, lifting F1-scores marginally to roughly $0.17$ at the expense of significantly increased computational time (taking between $69$ and $79$ seconds). As observed in previous datasets, the local Gemma LLM pipelines deliver the highest overall matching accuracy. The HB-LM-L and B25-LM-L configurations achieve peak F1-scores of approximately $0.49$ while maintaining robust recall rates above $0.8$. However, this superior accuracy incurs a tremendous computational burden, requiring over \num{2500} seconds to execute. The GPT-based matching models generally underperform their Gemma counterparts—--a trend consistent with our earlier observations on AMAZON-GOOGLE; for instance, BE-LM-R sees its recall drop significantly to $0.44$ and its F1-score fall to $0.4$, though it executes substantially faster at roughly $318$ seconds. For scenarios requiring practical efficiency, the canonical cross-matching methods (e.g., B25-CCM-L) offer a viable middle ground, yielding an F1-score of $0.23$ with an execution time of just $8.6$ seconds.

\begin{table}[ht]
\centering
\small
\caption{Evaluation results for AMAZON-WALMART.}
\label{tab:amazon_walmart_results}
\begin{tabular}{l c c c r r}
\toprule
\textbf{Method} & \textbf{Prec.} & \textbf{Rec.} & \textbf{F1} & \textbf{Time (sec)} & $\sigma$\\
\midrule
B25-B     & 0.077 & 0.859 & 0.141 & 1.42 &\\
B25-CM    & 0.092 & 0.855 & 0.167 & 69.21 &\\
B25-TS    & 0.090 & 0.787 & 0.161 & 49.85 &\\
B25-LM-L  & 0.348 & 0.806 & 0.486 & \num{2578.99} &15.62 \\
B25-LM-R  & 0.372 & 0.503 & 0.428 & 335.23 & 35.6\\
\midrule
BE-B      & 0.078 & 0.875 & 0.144 & 25.73 &\\
BE-CM     & 0.092 & 0.867 & 0.167 & 79.36 &\\
BE-TS     & 0.091 & 0.794 & 0.163 & 53.32 &\\
BE-LM-L   & 0.346 & 0.813 & 0.485 & \num{2635.47} &18.63 \\
BE-LM-R   & 0.362 & 0.440 & 0.397 & 318.73 &41.67\\
\midrule
HB-B      & 0.080 & 0.889 & 0.146 & 37.20 &\\
HB-CM     & 0.093 & 0.882 & 0.169 & 73.40 &\\
HB-TS     & 0.092 & 0.809 & 0.165 & 51.53 &\\
HB-LM-L   & 0.344 & 0.828 & 0.486 & \num{2579.28} &17.43\\
HB-LM-R   & 0.369 & 0.345 & 0.356 & \num{2089.66} &97.56\\
\midrule
B25-CB-L  & 0.081 & 0.904 & 0.149 & 0.20 &\\
B25-CCM-L & 0.130 & 0.883 & 0.227 & 8.59 &\\
B25-CTS-L & 0.124 & 0.829 & 0.216 & 9.62 &\\
\midrule
B25-CB-R  & 0.077 & 0.855 & 0.141 & 2.10 &\\
B25-CCM-R & 0.119 & 0.826 & 0.208 & 8.86 &\\
B25-CTS-R & 0.102 & 0.783 & 0.181 & 9.72 &\\
\bottomrule
\end{tabular}
\vspace{-5mm}
\end{table}

The \textbf{WDC-Products} (Table~\ref{tab:wdc_products_results}) benchmark highlights the severe scalability limitations of standard ER pipelines when applied to large datasets. The defining characteristic of this evaluation is the sheer computational time required. Even foundational baseline methods face substantial bottlenecks; B25-B requires over $512$ seconds, while BE-B escalates to nearly \num{2392} seconds. In order to elevate the recall to around $0.7$, we had to increase the retrieval parameter to $k=10$ for all architectures. Consequently, the blocking phase retrieved a total of $\vert{}A\vert{} \times 10 = 1\text{M}$ candidate pairs for resolution. Cross-encoder configurations (B25-CM, BE-CM, HB-CM) necessitate roughly \num{3695} to \num{3952} seconds to execute. Tensor search matching (e.g., B25-TS) offers a slight speedup over cross-encoders but still demands over \num{2700} seconds. Precision for all base blocking techniques plummets to roughly $0.03$, yielding minimal F1-scores confined between $0.06$ and $0.064$. While the cross-encoder matchers improve precision slightly to $0.1$, the highest achievable F1-score across all evaluated methods remains exceptionally low at just $0.186$ (B25-CM). 

This collapse in precision stems directly from two structural characteristics of the dataset. WDC-Products are saturated with distinct products sharing nearly identical descriptions—differing only by minor attributes like version numbers or capacities—which floods candidate pools with \textit{hard negatives} that baseline methods struggle to distinguish. Compounding this, the dataset features highly skewed matching cluster sizes; when applying a static top-$k$ retrieval strategy to entities with few true duplicates, the system inherently retrieves a massive volume of false positives, mathematically forcing the overall precision to collapse.

\begin{table}[ht]
\centering
\small
\caption{Evaluation results for WDC-Products.}
\label{tab:wdc_products_results}
\begin{tabular}{l c c c r}
\toprule
\textbf{Method} & \textbf{Prec.} & \textbf{Rec.} & \textbf{F1} & \textbf{Time (sec)} \\
\midrule
B25-B   & 0.034 & 0.715 & 0.064 & 512.35 \\
B25-CM  & 0.108 & 0.697 & 0.186 & \num{3695.04} \\
B25-TS  & 0.091 & 0.662 & 0.160 & \num{2732.18} \\
\midrule
BE-B    & 0.032 & 0.680 & 0.061 & \num{2391.84} \\
BE-CM   & 0.103 & 0.666 & 0.178 & \num{3952.48} \\
BE-TS   & 0.094 & 0.647 & 0.164 & \num{2718.90} \\
\midrule
HB-B    & 0.033 & 0.721 & 0.063 & \num{2675.73} \\
HB-CM   & 0.105 & 0.710 & 0.183 & \num{3863.10} \\
HB-TS   & 0.094 & 0.685 & 0.165 & \num{2738.13} \\
\bottomrule
\end{tabular}
\end{table}

\begin{table}[ht]
\centering
\small
\caption{Evaluation results for RETAILERS-C.}
\label{tab:retailers_c_results}
\begin{tabular}{l c c c r r}
\toprule
\textbf{Method} & \textbf{Prec.} & \textbf{Rec.} & \textbf{F1} & \textbf{Time (sec)} & $\sigma$\\
\midrule
B25-B     & 0.190 & 0.954 & 0.317 & 0.13 &\\
B25-CM    & 0.615 & 0.894 & 0.729 & 16.87 &\\
B25-TS    & 0.305 & 0.856 & 0.450 & 10.57 &\\
B25-LM-L  & 0.806 & 0.833 & 0.819 & \num{1356.47}&13.61\\
B25-LM-R  & 0.757 & 0.770 & 0.763 & 224.73 & 43.6\\
\midrule
BE-B      & 0.178 & 0.911 & 0.298 & 4.41 &\\
BE-CM     & 0.640 & 0.863 & 0.735 & 17.35 &\\
BE-TS     & 0.319 & 0.824 & 0.460 & 10.84 &\\
BE-LM-L   & 0.819 & 0.800 & 0.809 & \num{1273.68}& 11.21\\
BE-LM-R   & 0.853 & 0.768 & 0.808 & 139.50 & 32.52\\
\midrule
HB-B      & 0.187 & 0.958 & 0.313 & 5.11 &\\
HB-CM     & 0.616 & 0.901 & 0.732 & 17.34& \\
HB-TS     & 0.320 & 0.862 & 0.467 & 10.73 &\\
HB-LM-L   & 0.805 & 0.835 & 0.820 & \num{1277.8}& 12.12\\
HB-LM-R   & 0.863 & 0.788 & 0.824 & 159.84 & 28.43\\
\bottomrule
\end{tabular}
\vspace{-3mm}
\end{table}

The \textbf{RETAILERS-C} (Table~\ref{tab:retailers_c_results}) dataset yields notably higher baseline retrieval effectiveness and demonstrates the profound impact of advanced matching techniques. The blocking baselines achieve exceptionally high recall, ranging from $0.911$ to $0.958$. Unlike previous datasets where cross-encoders offered minimal gains, applying cross-encoder matching here (B25-CM, BE-CM, HB-CM) drives a massive boost in precision, leaping to over $0.61$. This translates into F1-scores between $0.729$ and $0.735$, all executing in highly efficient runtimes of roughly $17$ seconds. Conversely, tensor search matching (e.g., B25-TS, BE-TS) achieves far less impressive gains, topping out at an F1-score of $0.467$. The integration of LLMs delivers the highest overall accuracy, albeit with significant computational trade-offs. Both Gemma and GPT-based pipelines excel on this clean dataset. The Gemma models (B25-LM-L, BE-LM-L, HB-LM-L) consistently produce F1-scores between $0.809$ and $0.82$, maintaining robust recall above $0.8$, though they demand computationally heavy execution times exceeding \num{1270} seconds. Notably, the GPT matching configurations perform exceptionally well on this specific dataset. The HB-LM-R method achieves the peak overall F1-score of $0.824$ (driven by a high precision of $0.863$), and critically, it executes in just under $160$ seconds.

\begin{table}[ht]
\centering
\small
\caption{Evaluation results for RETAILERS-D.}
\label{tab:retailers_d_results}
\begin{tabular}{l c c c r r}
\toprule
\textbf{Method} & \textbf{Prec.} & \textbf{Rec.} & \textbf{F1} & \textbf{Time (sec)} & $\sigma$ \\
\midrule
B25-B     & 0.181 & 0.904 & 0.302 & 1.94& \\
B25-CM    & 0.427 & 0.868 & 0.572 & 138.19& \\
B25-TS    & 0.242 & 0.835 & 0.375 & 87.74& \\
B25-LM-L  & 0.559 & 0.808 & 0.661 & \num{11176.61}&46.12 \\
B25-LM-R  & 0.525 & 0.747 & 0.617 & \num{1851.52}&87.98 \\
\midrule
BE-B      & 0.179 & 0.894 & 0.298 & 23.74& \\
BE-CM     & 0.431 & 0.868 & 0.576 & 148.57& \\
BE-TS     & 0.253 & 0.828 & 0.388 & 89.9& \\
BE-LM-L   & 0.552 & 0.804 & 0.655 & \num{10451.43}&39.61 \\
BE-LM-R   & 0.575 & 0.772 & 0.659 & \num{1144.72}&82.62 \\
\midrule
HB-B      & 0.184 & 0.922 & 0.307 & 23.56& \\
HB-CM     & 0.416 & 0.888 & 0.567 & 149.36& \\
HB-TS     & 0.292 & 0.856 & 0.440 & 89.17& \\
HB-LM-L   & 0.544 & 0.823 & 0.655 & \num{10455.9}&31.31 \\
HB-LM-R   & 0.583 & 0.776 & 0.666 & \num{307.92}& 47.61\\
\bottomrule
\end{tabular}
\vspace{-3mm}
\end{table}

The \textbf{RETAILERS-D} (Table~\ref{tab:retailers_d_results}) benchmark introduces significant noise and structural irregularities, resulting in a noticeable degradation in overall matching effectiveness compared to its clean counterpart. Baseline blocking continues to preserve high recall rates (ranging from $0.89$ to $0.92$), but are bottlenecked by low precision (approximately $0.18$), trapping F1-scores near $0.30$. Applying cross-encoder verification (e.g., BE-CM) triggers a substantial improvement, nearly doubling the F1-score to $0.576$ with a precision of $0.431$, while executing in roughly $148$ seconds. Tensor search methods (e.g., B25-TS, BE-TS) fail to capture the same semantic nuance on the noisy text, underperforming cross-encoders with F1-scores confined to the $0.37$ to $0.44$ range. Once again, LLM-based matching pipelines dominate in accuracy, but highlight severe computational scalability issues. The local Gemma models (B25-LM-L, BE-LM-L, HB-LM-L) achieve robust F1-scores between $0.65$ and $0.66$, yet require a substantial amount of computational time—exceeding \num{10400} seconds. In stark contrast, the GPT-based configurations prove highly resilient and efficient on this dirty dataset. Specifically, HB-LM-R achieves the highest overall F1-score of $0.666$ with a strong precision of $0.583$. Crucially, it completes execution in approximately $307$ seconds. Compared to its hybrid Gemma counterpart (HB-LM-L), HB-LM-R delivers a slightly higher matching accuracy while executing nearly $34$ times faster. 

\subsection{Sensitivity, Stability, and Memory Footprint}

\begin{figure}[t] 	
 		\includegraphics[width=1\linewidth]{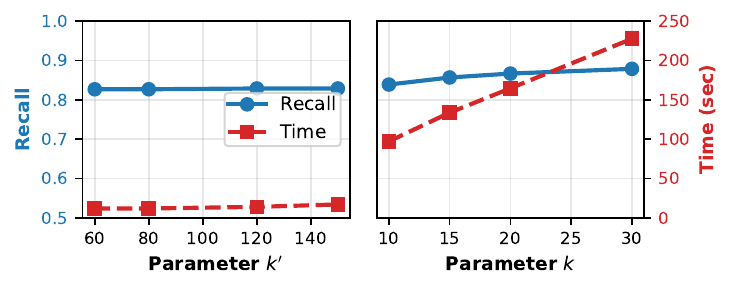}
		\caption{Trade-off between recall and execution time as a function of the retrieval parameters.}    
		\label{fig:k_ablation}	 
        \vspace{-3mm}
\end{figure}

\textbf{Parameters $k^{\prime}$ and $k$.} Across all evaluated pipelines, the final retrieval depth $k$ strictly bounds the number of candidates passed to the downstream matcher. In Hybrid Fusion Cascade, this is complemented by an intermediate depth $k^{\prime}$, which must be sufficiently large to pool candidates from both the lexical and semantic blockers. However, expanding these parameters introduces two distinct computational bottlenecks with severely diminishing returns in recall. As demonstrated by our isolated ablation measurements on the challenging AMAZON-GOOGLE dataset in Figure~\ref{fig:k_ablation}, increasing $k^{\prime}$ from $60$ to $150$ barely shifts blocking recall (from $0.827$ to $0.829$) while introducing a mild, yet unnecessary, CPU-bound overhead during RRF (from $11.69$ to $16.86$ seconds). Conversely, expanding the final candidate depth $k$ triggers a far more severe inference penalty, as it linearly inflates the total number of pairwise comparisons passed to the downstream matcher to exactly $k \cdot \vert{}A\vert{}$. Scaling $k$ from $10$ to $30$ yields a modest recall gain ($0.839$ to $0.879$) but forces the GPU-bound matching time to skyrocket from $97.1$ to $227.8$ seconds. Because both operations scale poorly against asymptotic recall improvements, they must be conservatively bounded.

\textbf{Variability of the Results.} The non-generative configurations---lexical, bi-encoder, tensor search, and cross-encoder models---exhibited near-deterministic behavior for both the resulting evaluation metrics ($\sigma \leq 0.02$) and end-to-end execution times. Conversely, for the generative LLM-based configurations, we explicitly enforced greedy decoding (\texttt{\allowbreak temperature = 0.0}) to suppress sampling noise and maximize reproducibility. Despite this strict parameterization, the local Gemma model still exhibited minor run-to-run fluctuations in inference latency and final evaluation metrics ($\sigma \leq 0.04$). We attribute this residual variance to the asynchronous execution of the out-of-process inference engine, dynamic KV-cache allocations,\footnote{The KV-cache stores intermediate attention tensors to prevent redundant computations during autoregressive generation. Allocating this GPU memory dynamically on the fly to accommodate varying output lengths introduces minor, non-deterministic latency overhead across batches.} and hardware-level CUDA non-determinism. In stark contrast, executions relying on the remote GPT API experienced significantly higher variability in end-to-end latency, nearly reaching a standard deviation of $\sigma \approx 100$ seconds on AMAZON-WALMART as illustrated in Table~\ref{tab:amazon_walmart_results}. This pronounced fluctuation is an expected artifact of network transmission overhead, dynamic server-side load balancing, and unpredictable queueing delays inherent to multi-tenant cloud endpoints.

\begin{table}[t]
\centering
\small
\caption{Peak Memory Utilization (GB) on WDC-Products.}
\label{tab:memory_utilization}
\begin{tabular}{llcc}
\toprule
\textbf{Blocking Method} & \textbf{Matching Method} & \textbf{CPU RAM} & \textbf{GPU VRAM} \\
\midrule
\multirow[t]{4}{*}{BM25} & -- & 4.10$^\dagger$ & --- \\
                                & Tensor Search & 4.72 & 0.42 \\
                                & Cross-Encoder & 4.23 & 0.49 \\
                                & Local LLM  & 4.21$^*$ & 8.75 \\
\midrule
\multirow[t]{4}{*}{Bi-Encoder} & -- & 14.50 & 0.42 \\
                                         & Tensor Search& 14.52 & 0.50 \\
                                         & Cross-Encoder& 14.52 & 1.16 \\
                                         & Local LLM & 14.52$^*$ & 8.72 \\
\midrule
\multirow[t]{4}{*}{Hybrid} & -- & 14.62 & 0.42 \\
                                      & Tensor Search  & 14.64 & 0.49 \\
                                      & Cross-Encoder  & 14.64 & 1.16 \\
                                      & Local LLM  & 14.64$^*$ & 8.72 \\
\bottomrule
\multicolumn{4}{l}{\footnotesize $^\dagger$Tantivy index consumes $0.19$GB; remaining RAM reflects the ER data structures.}\\
\multicolumn{4}{l}{\footnotesize $^*$LLM executed via Ollama server; CPU RAM solely reflects the ER data structures.}
\end{tabular}
\vspace{-5mm}
\end{table}

\textbf{Memory Footprint.} Table~\ref{tab:memory_utilization} summarizes the peak host memory (CPU RAM) and device memory (GPU VRAM) across all blocking and matching combinations on the WDC-Products benchmark. Transitioning from lexical blocking to dense ANNS, using an HNSW index, incurs a substantial CPU memory expansion from $4.10$~GB to $14.50$~GB (a $3.5\times$ increase). While Tantivy for BM25 maintains an exceptionally compact index footprint of $0.19$~GB, the HNSW graph must co-locate raw dense float vectors alongside multi-layer spatial adjacency lists in uncompressed host memory. Furthermore, the blocking phase establishes the absolute high-water mark for host RAM; downstream matching cascades (Tensor Search and Cross-Encoder) do not inflate CPU memory further ($14.52\text{--}14.64$~GB across all dense and hybrid runs). Conversely, GPU VRAM remains modest for the cross-encoder and tensor search ($0.42\text{--}1.16$~GB) because matching operates in batched, pairwise inference mode under FP16 precision without maintaining computational gradient graphs. Consequently, even multi-stage neural cascades can be deployed on standard commodity GPUs, while scaling to larger entity corpora requires addressing the CPU memory footprint of the ANNS index.

Substituting these lightweight matchers with a local generative LLM requires careful infrastructural provisioning to remain viable on local hardware. To evaluate this, the specific Gemma model used, with its 12B parameters, was executed via the Ollama backend utilizing 4-bit quantization\footnote{4-bit quantization compresses model weights from standard half-precision (FP16) down to 4-bit integers, significantly reducing static VRAM requirements with negligible accuracy loss.} and KV cache. This optimization strategy successfully reduced the model's peak VRAM footprint from $24$~GB down to an observed peak usage of $8.72$~GB. Furthermore, because Ollama operates as an isolated daemon process, the host application accessed the model exclusively via stateless HTTP requests. Because Ollama executes as a separate process, its model-memory footprint is not reflected in the Python process's resident memory measurements.

To mitigate the memory overhead of the dense index, we applied product quantization~\cite{pq} configured with 96 subquantizers. This configuration compressed the raw vector storage footprint from $2.86$~GB down to approximately $0.18$~GB, reducing the total index size  from $14.5$ GB to roughly $11.8$~GB. However, this compression resulted in a recall reduction of nearly $4\%$.

\subsection{Discussion}

The structural evaluation of the blocking methods reveals a strict trade-off between architectural complexity, computational overhead, and semantic accuracy:
\begin{itemize}[leftmargin=*]
    \item Bi-encoders excel on verbose, attribute-rich data (e.g., ABT-BUY and AMAZON-GOOGLE), while BM25 dominates on concise, exact-match retail titles (eg., RETAILERS). Hybrid Fusion consistently achieves the highest or near-highest blocking recall across the evaluated datasets.
    \item Hybrid Fusion attempts to combine exact lexical anchoring with semantic retrieval in  order to resolve the blind spots of both isolated paradigms. The experimental evaluation generally supports this synergy across diverse data distributions, yielding modest but consistent improvements: on the AMAZON-WALMART benchmark, Hybrid blocking achieves a recall of $0.889$, surpassing both the standalone lexical ($0.859$) and semantic ($0.875$) baselines. Similarly, on RETAILERS-C and RETAILERS-D, the dual-retrieval stream drives recall to peak values of $0.958$ and $0.922$, respectively.
    \item BM25 processes massive catalogs in a fraction of the time required by bi-encoders, which suffer from both the linear time vectorization process and HNSW graph traversal costs. This efficiency becomes particularly pronounced on the massive WDC-Products benchmark, where BM25 operates approximately $4.7$ times faster than the bi-encoder with HNSW, executing in just $512.35$ seconds compared to \num{2391.84} seconds. 
    \item BM25 also maintains an exceptionally compact index footprint ($0.19$~GB index overhead, $4.23$~GB total RAM on WDC-Products), standing in stark contrast to the bi-encoder's HNSW index, which inflates host memory to $14.50$~GB. This difference makes BM25 the most scalable baseline when computational or memory constraints are severe.  
\end{itemize}

The experimental evaluation of the downstream matching stage establishes a strict architectural hierarchy, governed by the tension between computational latency and semantic accuracy:
\vspace{-0.2cm}
\begin{itemize}[leftmargin=*]
    \item For small to moderately sized datasets, the cross-encoder provides the most reliable and cost-effective leap in precision across all evaluated pipelines. However, as demonstrated by the WDC-Products benchmark, its inference cost becomes a severe computational bottleneck at extreme scales unless paired with aggressive candidate filtering in the blocking phase.
    \item Tensor search matching consistently underperforms full cross-encoders across all evaluation metrics, experiencing severe precision degradation on the noisy RETAILERS benchmarks.
    \item The execution speedup that tensor search offers over cross-encoders (e.g., \num{2732.18} seconds vs. \num{3695.04} seconds on WDC-Products within the lexical cascade) fails to compensate for the significant drop in overall F1-score.
    \item Generative verification provides a strong zero-shot matching baseline, while exposing a stark divergence between local and remote models. The local quantized Gemma model achieves the highest F1-scores on standard English datasets but suffers from prohibitive inference latency, requiring over \num{10400} seconds on RETAILERS-D. 
    \item Conversely, the remote GPT model excels on RETAILERS by leveraging its vast parametric knowledge to resolve heterogeneous promotional noise and inconsistent measurement units. 
    \item Highlighting this efficiency, the hybrid GPT configuration (HB-LM-R) achieves the peak F1-score of $0.666$ on RETAILERS-D while executing roughly $34$ times faster than its local Gemma counterpart.
    \item In stark contrast to the RETAILERS benchmarks—which primarily comprises concise, high-signal product titles—the remote GPT pipelines (B25-LM-R, BE-LM-R, and HB-LM-R) suffer a severe recall drop on ABT-BUY, AMAZON-GOOGLE, and AMAZON-WALMART. This degradation is directly attributed to the verbose nature of these datasets. The lengthy, unstructured descriptions introduce substantial textual noise, a problem exacerbated by the frequent absence of explicit model numbers in many records. This vulnerability is particularly pronounced in the AMAZON-GOOGLE dataset, where, deprived of these strong identifying anchors to cut through the context, the recall rates for B25-LM-R, BE-LM-R, and HB-LM-R plummet to just $0.259$, $0.143$, and $0.163$, respectively. 
\end{itemize}

Results show no single cascade architecture dominates; instead, each presents trade-offs across vocabulary sensitivity, geometric representation density, latency, and memory consumption:

\begin{itemize}[leftmargin=*]
    \item The Lexical Cascade establishes a highly competitive accuracy ceiling, making the adoption of computationally heavy semantic models difficult to justify in resource-constrained environments. On datasets like RETAILERS, lexical methods (B25-CM) deliver matching quality nearly identical to dense vector pipelines (e.g., $0.729$ vs. $0.735$ on RETAILERS-C, and $0.572$ vs. $0.576$ on RETAILERS-D). Furthermore, because BM25 is so efficient, the pipeline's total execution time is almost entirely dominated by the downstream matching stage. On ABT-BUY, for instance, while B25-CM and BE-CM require similar total execution times ($15.40$ seconds vs. $15.09$ seconds), their internal latency profiles differ drastically: lexical blocking consumes a mere $0.17$ seconds, whereas  semantic blocking requires $4.99$ seconds of the pipeline.  
    \item While the Semantic Cascade excels at retrieving broad conceptual matches (e.g., achieving $0.944$ blocking recall on ABT-BUY), dense vectors blur fine-grained alphanumeric distinctions. This floods the candidate pool with topologically adjacent \textit{hard negatives}, keeping F1-scores of BE-CM and BE-TS trapped near $0.32$ despite the high initial recall. Only LLM-based matching filters these \textit{hard negatives} to restore high precision (e.g., $0.833$ and $0.726$ on ABT-BUY; $0.819$ and $0.853$ on RETAILERS-C for BE-LM-L and BE-LM-R).
    \item The Hybrid Fusion pipelines present a strict infrastructural trade-off: they provide a memory-free, yet ultimately marginal, accuracy upgrade at the cost of severe CPU latency. While adding discrete BM25 retrieval to an existing semantic pipeline incurs virtually zero additional RAM overhead (e.g., adding just $120$~MB, a $+0.83\%$ increase, on WDC-Products), the resulting performance gains are often negligible. For example, cross-encoder F1-scores stagnate near $0.28$ on AMAZON-GOOGLE for B25-CM, BE-CM, and HB-CM, and Hybrid Fusion even trails the semantic baseline slightly on ABT-BUY ($0.322$ vs. $0.323$). Maintaining dual indexes and merging redundant candidate lists for such minimal returns forces the blocking phase to execute up to $48\times$ slower than a standard Lexical Cascade (e.g., $32.09$ seconds versus $0.67$ seconds on AMAZON-GOOGLE).    
    \item The Generative-Lexical Cascade delivers an extreme operational speedup in exchange for a modest reduction in accuracy compared to local LLM pairwise verification. This trade-off is clearly demonstrated on AMAZON-GOOGLE, where canonical matching via GPT (B25-CCM-R) yields an F1-score of $0.432$ (with $0.788$ recall) in just $3.67$ seconds. In comparison, online pairwise verification via local Gemma (BE-LM-L) requires \num{2732.63} seconds to reach an F1-score of $0.527$, confirming a $>700\times$ acceleration for the cascade. 
    \item While canonicalization in the Generative-Lexical cascade eliminates BM25's vocabulary fragility without dense indexing overhead, it introduces upfront offline GPU compute demands, which are linear to the size of the dataset, and risks error propagation through generative hallucinations.
\end{itemize}

To systematically isolate the structural vulnerabilities of discrete and continuous retrieval, we conducted a rigorous separability analysis, with the complete methodology and visualizations detailed in Appendix~\ref{appendix:eparability}. Separability quantifies the capacity of a given feature space to distinctly partition true matching entities from structurally similar non-matching entities. Rather than evaluating trivial random cross-joins, we constructed a highly challenging evaluation corpus by actively mining hard negatives. By plotting the continuous probability distributions of these scores via Kernel Density Estimation, we visualized the exact density overlap between true matches and hard negatives across both lexical and semantic spaces. This separability evaluation directly underpins the empirical behavior of the evaluated end-to-end cascades:
\vspace{-0.2cm}
\begin{itemize}[leftmargin=*]
    \item The density estimations reveal that while true matches cluster tightly near a normalized semantic similarity of $1.0$, hard negatives heavily skew toward the $[0.7, 0.95]$ range due to shared domain terminology and brand vocabulary. This geometric overlap directly explains why standalone semantic blocking (BE-B) achieves poor precision (e.g., $0.032$ on WDC-Products, $0.078$ on AMAZON-WALMART, or $0.135$ on AMAZON-GOOGLE) and necessitates deep cross-encoder verification to untangle dense topological neighborhoods where cosine similarity alone fails.  
    \item In the lexical similarity distributions, non-matches concentrate near zero, but the true match distributions exhibit a long left tail. These low-scoring true matches represent pairs suffering from disjoint vocabularies or severe formatting differences. When BM25 blocking enforces a strict top-$k$ candidate threshold, items in this low-similarity tail are permanently discarded, causing the noticeable drop in lexical blocking recall on noisy datasets (e.g., $0.715$ on WDC-Products).  
    \item The separability analysis also demonstrates that the failure modes of lexical and semantic retrieval are substantially complementary. Lexical metrics isolate non-matches near zero but fail on true matches with lexical variation, whereas semantic metrics reliably pull true matches toward $1.0$ but inadvertently drag hard negatives with them. This distribution divergence directly justifies the synergy of RRF, mathematically explaining the peak blocking recall observed across the Hybrid Fusion benchmarks. 
\end{itemize}

\section{Conclusions and Future Work}
\label{sec:conclusions}
We presented an end-to-end evaluation of PER pipelines, examining the trade-offs among blocking recall, matching accuracy, computational cost, and infrastructure requirements. Our results show that no single architecture consistently dominates across datasets.

Lexical blocking is preferable when efficiency and precise vocabulary matching are priorities, while semantic retrieval becomes valuable when lexical variation is substantial. At the matching stage, cross-encoders emerge as a practical default for balancing effectiveness and cost, whereas tensor search provides insufficient accuracy gains to justify its use in our experiments. LLM-based verification is better suited to accuracy-critical or smaller-scale settings where its computational and financial cost can be tolerated. Thus, the appropriate PER architecture ultimately depends on the characteristics of the data and the computational constraints of the deployment environment.

As future work, we will focus on integrating disk-aware ANN indexes, such as DiskANN~\cite{diskann}, and developing adaptive blocking strategies that dynamically balance lexical and semantic retrieval while reducing the overhead of the hybrid architecture.

\balance
\bibliographystyle{ACM-Reference-Format}
\bibliography{references}
\newpage
\appendix
\section{Hybrid Candidate Retrieval and Reciprocal Rank Fusion}
\label{appendix:hybrid_rrf}
For each query $q$, we first retrieve the top-$k^\prime$ candidates independently from the BM25 lexical index and the ANNS dense vector index. To streamline retrieval efficiency, all vector embeddings are normalized to unit length, enabling direct inner product computations that are mathematically equivalent to cosine similarity. To unify the disparate score distributions of BM25 (unbounded positive real values) and inner product similarities (bounded within $[-1, 1]$), we merge these retrieved sets into a joint candidate pool $\mathcal{D} = \mathcal{D}_{\text{lex}} \cup \mathcal{D}_{\text{sem}}$. Specifically, $\mathcal{D}_{\text{lex}}$ represents the top-$k$ candidate records returned by the lexical retrieval method (BM25 index), while $\mathcal{D}_{\text{sem}}$ comprises the candidates retrieved via the semantic retrieval method (ANNS dense vector index).

For each candidate $c$, a fused relevance score is calculated using position-based Reciprocal Rank Fusion~\cite{rrf}:
\begin{equation}
    \text{RRF}(c) = \sum_{m \in \{\text{lex}, \text{sem}\}} \mathbb{I}(c \in \mathcal{D}_m) \cdot \frac{1}{\mathcal{G} + r_m(c)}
\end{equation}

Within this set, $r_m \in \{0, 1, \dots, k^\prime - 1\}$ denotes the zero-indexed ordinal rank assigned to candidate $c$ by retrieval method $m$ (lexical or semantic). Smoothing constant $\mathcal{G} = 60$ attenuates the dominance of high-ranked items, and $\mathbb{I}(\cdot)$ is an indicator function defined as:
\begin{equation}
    \mathbb{I}(c \in \mathcal{D}_m) = 
    \begin{cases} 
      1 & \text{if } c \in \mathcal{D}_m \\ 
      0 & \text{otherwise} 
    \end{cases}
\end{equation}
ensuring that reciprocal scoring is evaluated exclusively for candidates present in retrieval method $m$'s candidate pool. The final top-$k$ retrieved results $\mathcal{R}_k$ correspond to the $k$ elements in $\mathcal{D}$ possessing the highest aggregated $\text{RRF}$ scores.

To illustrate the practical efficacy of Reciprocal Rank Fusion, consider a retrieval scenario where a user searches for an exact product identifier, such as the query \textit{``H-10AM5''}. The system evaluates two candidates: $c_1$ (\textit{``Sony H-10AM5 Headphones''}), which represents the exact target, and $c_2$ (\textit{``Sony H-10AM9 Headphones''}), a semantically similar but factually distinct item. In the lexical stream, BM25 natively prioritizes exact token matches; thus, $c_1$ achieves the top rank ($r_{\text{lex}} = 0$), while $c_2$ is severely penalized for missing the exact identifier, falling to a significantly lower position, such as $r_{\text{lex}} = 49$. Conversely, the dense embedding model perceives both strings as structurally identical domain concepts, occasionally ranking the semantic near-match higher. Assuming $c_2$ secures the top semantic rank ($r_{\text{sem}} = 0$) and $c_1$ follows closely at $r_{\text{sem}} = 3$, the RRF mechanism effectively resolves this conflict. Utilizing the standard smoothing constant $\mathcal{G} = 60$, the final fused score for $c_1$ is calculated as $\frac{1}{60 + 0} + \frac{1}{60 + 3} \approx 0.0325$. Meanwhile, $c_2$ yields a score of $\frac{1}{60 + 49} + \frac{1}{60 + 0} \approx 0.0258$. Because $c_1$ secures the maximum possible RRF fraction from the lexical stream while maintaining a competitive semantic rank, its compounded score easily surpasses $c_2$. This demonstrates how a top BM25 rank acts as an algorithmic anchor, mathematically safeguarding exact lexical matches from being overridden by fuzzy semantic approximations.

Conversely, RRF is equally critical for elevating candidates that capture the true conceptual intent of a query despite lacking exact keyword overlap. Consider a scenario where a user searches for the query \textit{``noise cancelling wireless earbuds''}. The system evaluates candidate $c_1$ (\textit{``True Bluetooth in-ear headphones with active sound isolation''}), which represents the exact product the user intends to find, alongside candidate $c_2$ (\textit{``Replacement silicone tips for noise cancelling wireless earbuds''}), which is merely a cheap accessory. Because $c_2$ contains an exact token-for-token match with the query, the BM25 lexical stream erroneously ranks it first. Candidate $c_1$, missing the specific tokens \textit{``cancelling''} and \textit{``earbuds''}, is penalized by BM25 and falls to a lower position, such as $r_{\text{lex}} = 15$. However, the dense embedding model successfully maps the synonymous concepts between the query and $c_1$, recognizing it as a primary electronic device and granting it the top semantic rank ($r_{\text{sem}} = 0$). The model correctly identifies candidate B as a structurally distinct accessory, dropping it to $r_{\text{sem}} = 45$. Applying RRF, the conceptually aligned $c_1$ achieves a fused score of $\frac{1}{60 + 15} + \frac{1}{60 + 0} \approx 0.0300$. In contrast, the lexically over-fitted $c_2$ yields a lower combined score of $\frac{1}{60 + 0} + \frac{1}{60 + 45} \approx 0.0262$. In this instance, RRF mathematically suppresses the keyword-stuffed accessory, allowing the semantically superior document to emerge as the final top-ranked result.

\section{Lexical and Semantic Separability}
\label{appendix:eparability}
In the context of PER, \textit{separability} refers to the capacity of a given feature space—whether lexical or semantic—to distinctly partition true matching entities from non-matching entities. Visualizing this separability explicitly quantifies the density overlap between true matches and highly similar distinct items (\textit{hard negatives}), underscoring the profound complexity of the PER task. To rigorously evaluate the separability of the PER pipelines, we construct an evaluation corpus by mining \textit{hard negatives} rather than relying on trivial random cross-joins. Specifically, we uniformly sample 10\% of records from $A$ (capped at \num{5000} records for computational tractability). For each such anchor record, we employ a \textit{retrieve-and-filter} approach that extracts the top-$5$ nearest neighbors from source $B$. 

\begin{figure}[h] 	
 		\includegraphics[width=1\linewidth]{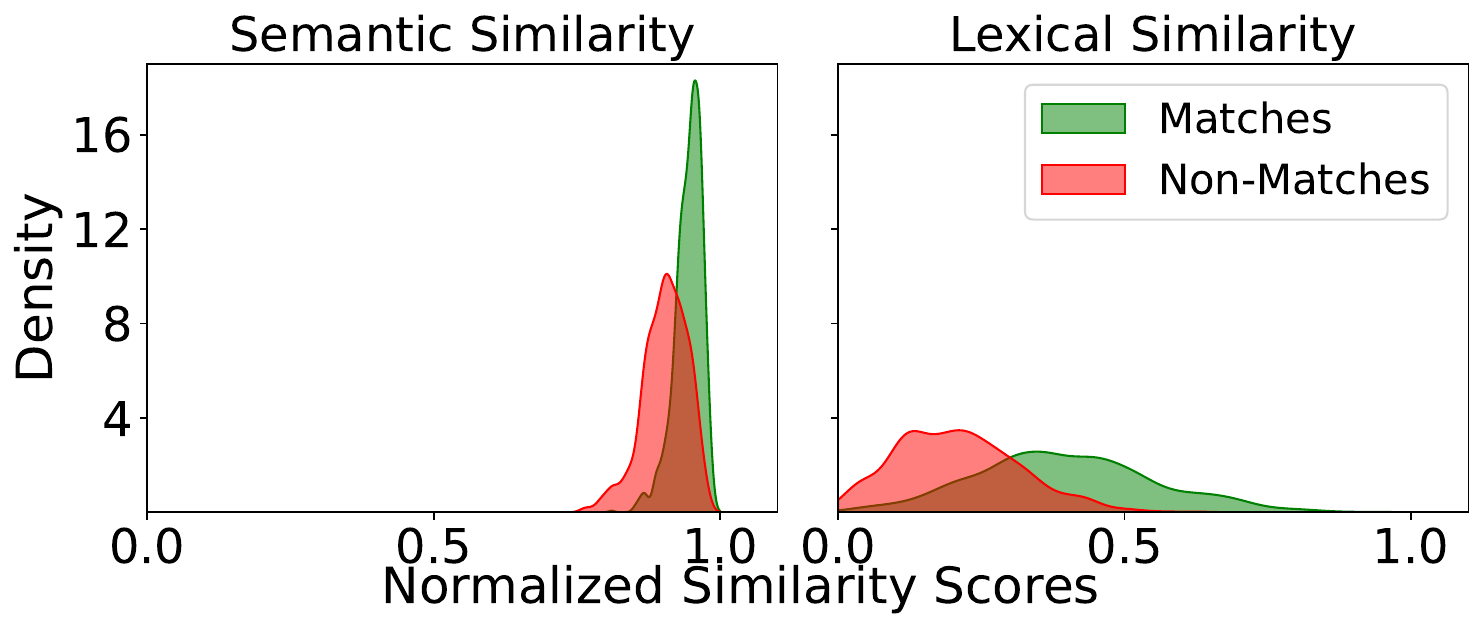}
		\caption{ABT-BUY separability}    
		\label{fig:abt_buy_separability}	 
        \vspace{-2mm}
\end{figure}
\begin{figure}[h] 	
 		\includegraphics[width=1\linewidth]{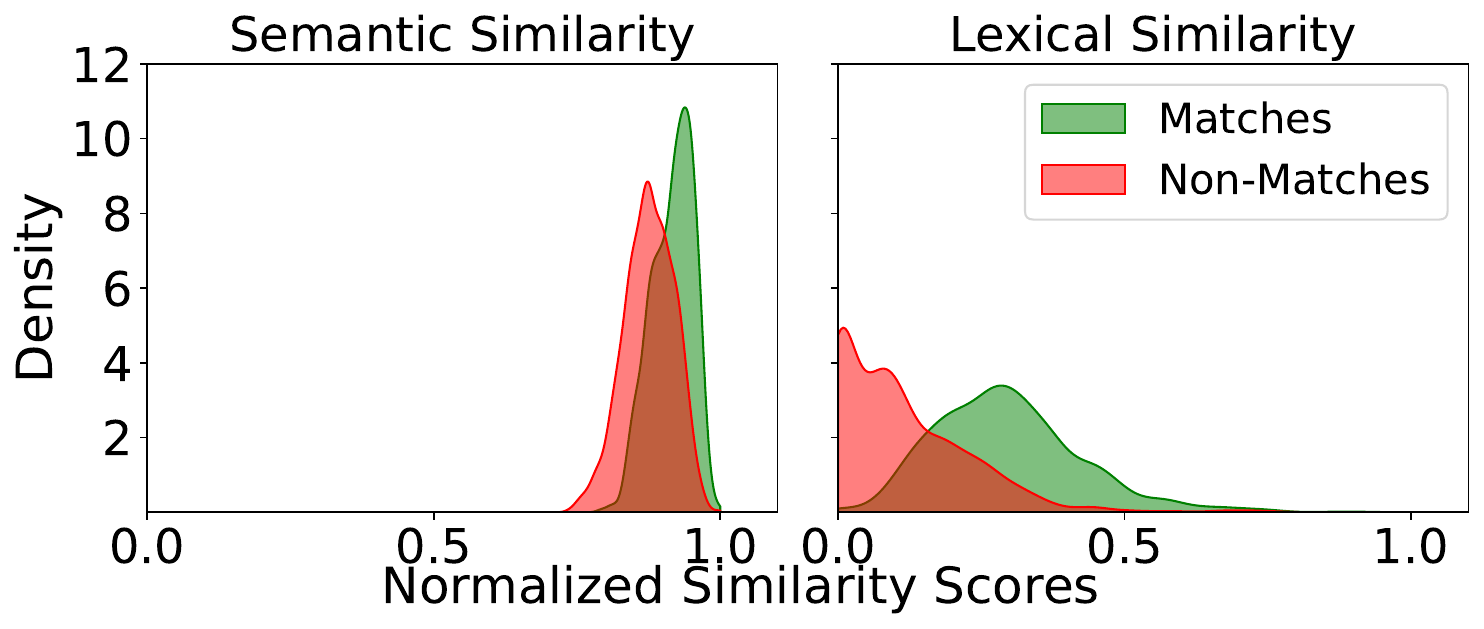}
		\caption{AMAZON-GOOGLE separability}    
		\label{fig:amazon-google_separability}	 
        \vspace{-2mm}
\end{figure}

This retrieval is performed using both exact cosine similarities over the dense vector embeddings, and lexical token-matching via BM25, subsequently filtering out any known ground-truth matches. This methodology yields a highly challenging candidate pool with a controlled positive-to-hard-negative ratio ranging between 1:1 and 1:3.\footnote{The exact ratio fluctuates per dataset due to three factors: (1) the deduplication of overlapping candidates retrieved by both the semantic and lexical methods; (2) the subtraction of true positive matches, which frequently appear in the top-$5$ results; and (3) the fixed \num{5000} anchor sampling cap, which scales the total volume of generated negatives differently relative to the fixed total number of ground-truth positives in each specific benchmark.} We then utilize Kernel Density Estimation plots to visualize the continuous probability distributions of normalized semantic and lexical similarities across both the match and non-match populations. 

\begin{figure}[h] 	
 		\includegraphics[width=1\linewidth]{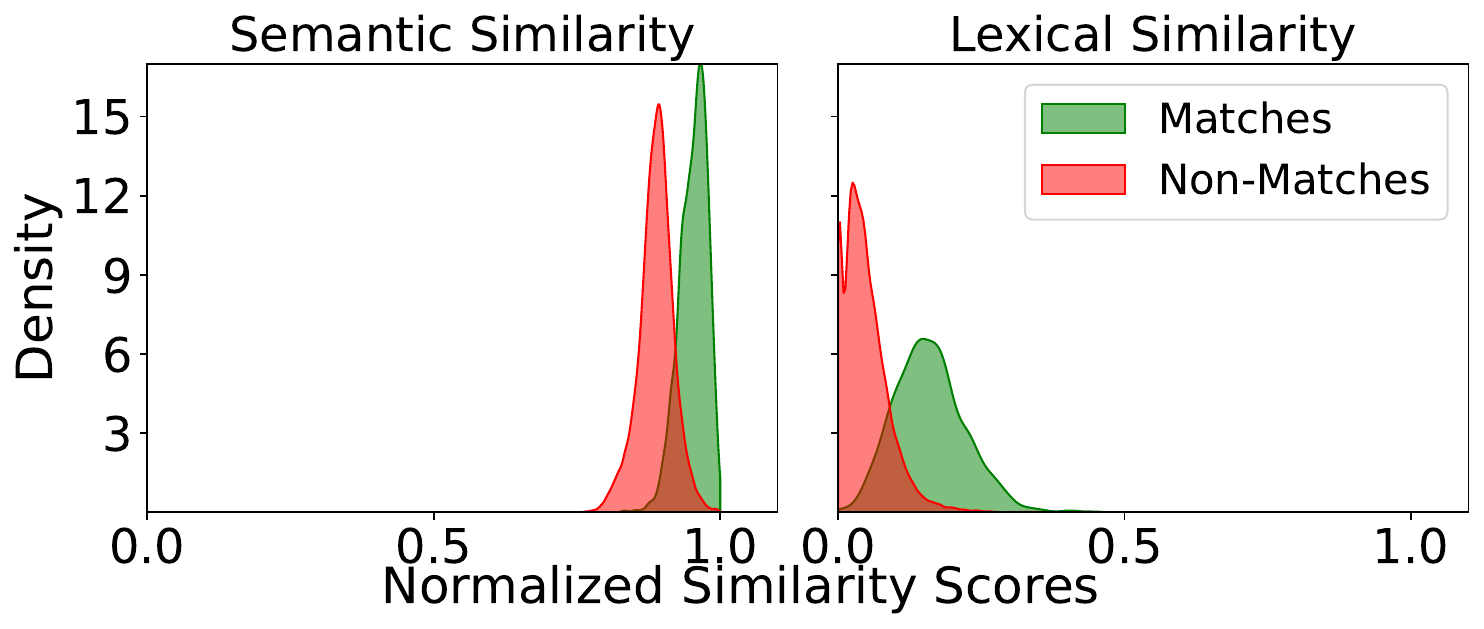}
		\caption{AMAZON-WALMART separability}    
		\label{fig:amazon-walmart_separability}	 
        \vspace{-2mm}
\end{figure}
\begin{figure}[h] 	
 		\includegraphics[width=1\linewidth]{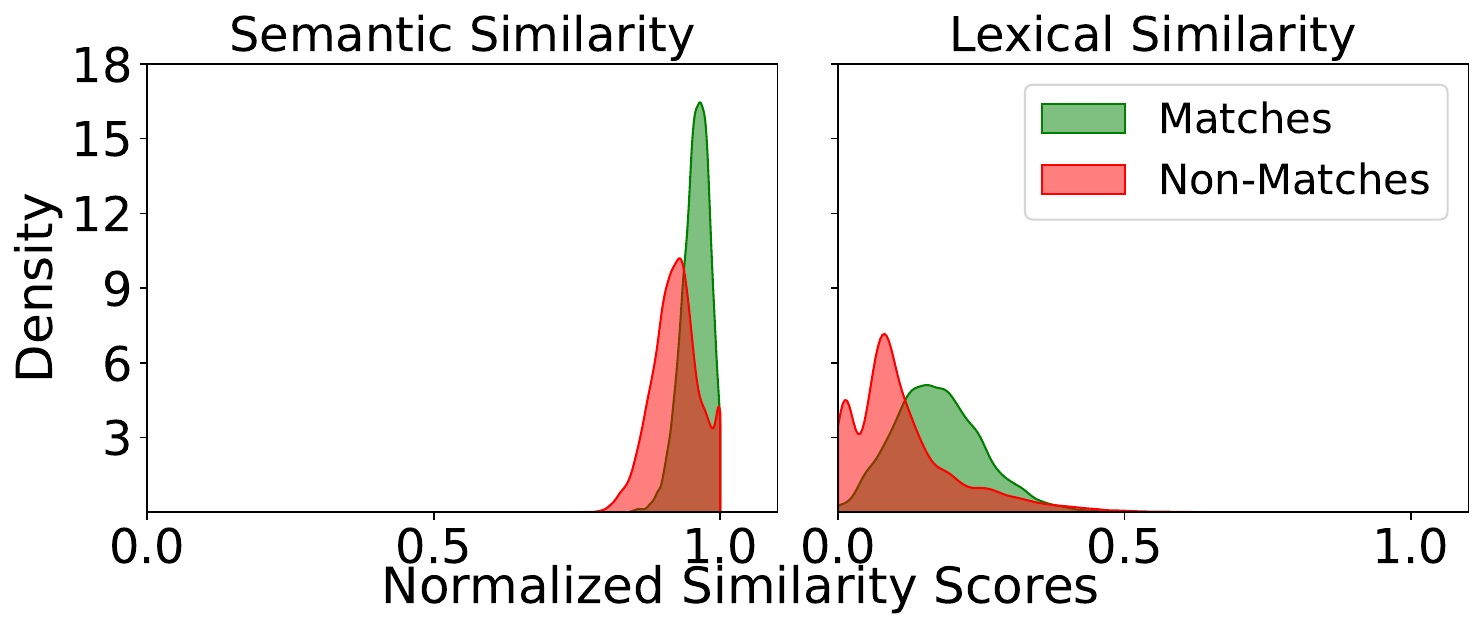}
		\caption{WDC-Products separability}    
		\label{fig:wdc_products_separability}	 
        \vspace{-2mm}
\end{figure}

\begin{figure}[h] 	
 		\includegraphics[width=1\linewidth]{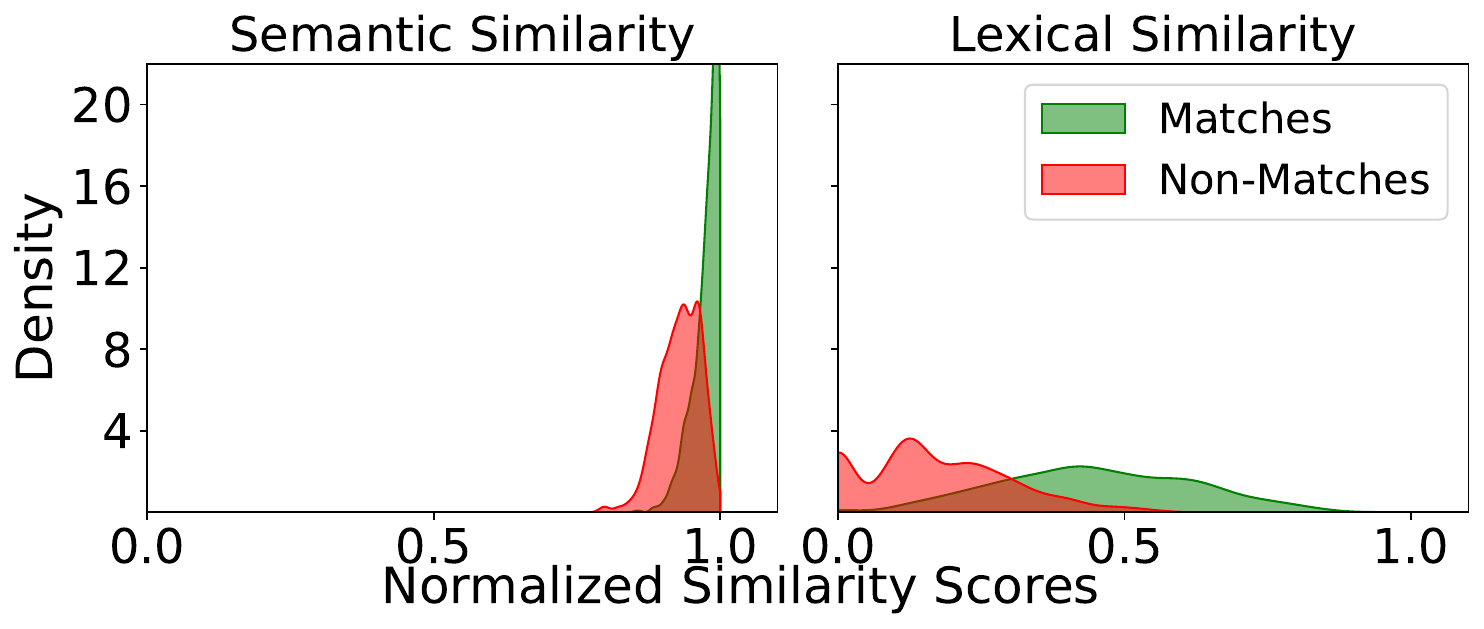}
		\caption{RETAILERS-D separability}    
		\label{fig:retailers_separability}	 
        \vspace{-2mm}
\end{figure}

The separability analysis reveals distinct behavioral differences between the semantic and lexical feature spaces (ABT-BUY in Figure~\ref{fig:abt_buy_separability}, AMAZON-GOOGLE in Figure~\ref{fig:amazon-google_separability}, AMAZON-WALMART in Figure~\ref{fig:amazon-walmart_separability}, WDC-Products in Figure~\ref{fig:wdc_products_separability}, and RETAILERS-D in Figure \ref{fig:retailers_separability} ), the lexical similarity distributions demonstrate a moderate degree of separability; because highly discriminative attributes like product titles carry the true signal, token-based metrics  isolate non-matches, concentrating their density heavily between $0.0$ and $0.25$. Conversely, the semantic similarity distributions illustrate the recurring vulnerability of dense vector spaces to shared domain terminology. While true matches cluster tightly near the $1.0$ threshold, explicitly mined hard negatives—such as distinct products from the same brand—share a dense technical vocabulary. This causes the similarity scores of non-matches to shift toward the upper end of the $[0.5, 1.0]$ interval, creating a significant region of density overlap (with the notable exception of the AMAZON-WALMART dataset) that complicates any decision boundary. This overlap indicates that neither representation provides a consistently clean decision boundary across the evaluated datasets.

\end{document}